\documentclass{article}
\usepackage{ijcai21}

\usepackage{times}

\usepackage{soul}
\usepackage{url}
\usepackage[hidelinks]{hyperref}
\usepackage[utf8]{inputenc}
\usepackage[small]{caption}
\usepackage{graphicx}
\usepackage{amsmath}
\usepackage{booktabs}

\usepackage{paralist} 
\usepackage[noend,ruled]{algorithm2e} 
\usepackage{amssymb} 
\usepackage{subcaption} 
\usepackage{listings} 
\usepackage{xcolor} 

\definecolor{codegreen}{rgb}{0.47,0.66,.37}
\definecolor{codegray}{rgb}{0.95,0.95,0.95}
\definecolor{codepurple}{rgb}{0.53,0,0}
\definecolor{backcolour}{rgb}{.96, .96, .96}
\definecolor{textcolor}{rgb}{0, 0, 0}
\definecolor{functioncolor}{rgb}{0.22, 0.45, 0}
\definecolor{keywordcolor}{rgb}{0, 0, 0.8}
 
\lstdefinestyle{mystyle}{
    backgroundcolor=\color{backcolour},   
    commentstyle=\color{codegreen},
    keywordstyle=\color{keywordcolor},
    numberstyle=\tiny\color{textcolor},
    stringstyle=\color{codepurple},
    emphstyle=\color{functioncolor}
    basicstyle=\ttfamily\footnotesize\color{textcolor},
    breakatwhitespace=false,         
    breaklines=true,                 
    captionpos=t,                    
    keepspaces=false,                 
    numbersep=2pt,                  
    showspaces=false,                
    showstringspaces=false,
    showtabs=false,                  
    tabsize=2
}
 
\newtheorem{THEOREM}{Theorem}
\newenvironment{Theorem}{\begin{THEOREM} \hspace{-.85em} {\bf .}}{\end{THEOREM}}
\newtheorem{DEFINITION}[THEOREM]{Definition}
\newenvironment{definition}{\begin{DEFINITION} \hspace{-.85em} {\bf .} \rm}{\end{DEFINITION}}
\newtheorem{EXAMPLE}[THEOREM]{Example}
\newenvironment{example}{\begin{EXAMPLE} \hspace{-.85em} {\bf .} \rm}{\end{EXAMPLE}}
\newenvironment{proof}{\noindent {\bf Proof.} }{}
\newtheorem{PROPOSITION}[THEOREM]{Proposition}
\newenvironment{proposition}{\begin{PROPOSITION} \hspace{-.85em} {\bf .} \rm}{\end{PROPOSITION}}
\newcommand{\bbox}{\vrule height7pt width4pt depth1pt}
\newcommand{\qed}{\bbox\vspace{0.1in}}  

\title{Learning Implicitly with Noisy Data in Linear Arithmetic}

\author{
Alexander P Rader$^1$\and
Ionela G Mocanu $^2$\and
Vaishak Belle $^2$\And
Brendan Juba $^3$\\
\affiliations
$^1$Imperial College London\\
$^2$University of Edinburgh\\
$^3$Washington University in St. Louis\\
\emails
alexander.rader20@imperial.ac.uk,
\{i.g.mocanu, vaishak\}@ed.ac.uk,
bjuba@wustl.edu
}

\begin{document}
\maketitle

\begin{abstract}
Robust learning in expressive languages with real-world data continues to be a challenging task. Numerous conventional methods appeal to heuristics without any assurances of robustness. While probably approximately correct (PAC) Semantics offers strong guarantees, learning explicit representations is not tractable, even in propositional logic. However, recent work on so-called “implicit" learning has shown tremendous promise in terms of obtaining polynomial-time results for fragments of first-order logic. In this work, we extend implicit learning in PAC-Semantics to handle noisy data in the form of intervals and threshold uncertainty in the language of linear arithmetic. We prove that our extended framework keeps the existing polynomial-time complexity guarantees. Furthermore, we provide the first empirical investigation of this hitherto purely theoretical framework. Using benchmark problems, we show that our implicit approach to learning optimal linear programming objective constraints significantly outperforms an explicit approach in practice.\footnote{This is an extended version of our IJCAI21 paper \cite{rader2021noisy}}
\end{abstract}

\section{Introduction}
Data in the real world can be incomplete, noisy and imprecise. Approaches from the knowledge representation communities take great care to represent expert knowledge; however, this knowledge can be hard to come by, challenging to formalize for non-experts, and brittle, especially in the presence of noisy measurements. In contrast, connectionist approaches, such as neural networks, have been particularly successful in learning from real-world data. However, they represent knowledge as distributed networks of nodes, rendering it difficult to incorporate knowledge from other sources. Moreover, such methods usually do not come with any guarantees of soundness, and are also quite brittle to small shifts in the domain \cite{pmlr-v97-recht19a,koh2020wilds}.

In this work, we are concerned with learning in expressive languages, such as fragments of first-order logic, where knowledge and queries are represented as logical formulas. In that regard, Valiant \shortcite{Valiant2000} recognized that the challenge of learning should be integrated with deduction. In particular, he proposed a semantics to capture the quality possessed by the output of  PAC-learning algorithms when formulated in a logic. 

Unfortunately, many results on learning logical languages have been discouraging. For example, in agnostic learning, \cite{kearns1994toward},
where one does not require examples to be fully consistent with learned sentences, 
efficient algorithms for learning conjunctions of formulas would yield an efficient algorithm for PAC-learning disjunctive normal forms (DNF), which current evidence suggests to be intractable  \cite{daniely2016complexity}.  

Interestingly, {Khardon and Roth} \shortcite{KhardonRothJournal1997} and {Juba} \shortcite{juba2013} observed that, by circumventing the need to produce an explicit representation, learning to reason can be effectively reduced to classical reasoning, leading to a notion of ``implicit learning'' Since reasoning in propositional logic is NP-complete, implicit learning in full propositional logic also cannot be done in polynomial time (unless P=NP), but when limited to tractable fragments of propositional logic, learning inherits the tractability guarantee. Very recently, the learnability results have been extended to first-order clauses in \cite{bellejuba2019}, and then to certain fragments of \textit{satisfiability modulo theories} (SMT) in \cite{PAC+SMT}.

In this work, we extend the implicit learning approach to handle partial information given by bounds (e.g., intervals), thus capturing an even larger class of problem settings, and develop an implementation strategy for a hitherto purely theoretical framework. 

\subsection{Theoretical Contributions}
\begin{compactenum}
\item Extending the PAC-Semantics framework to be able to handle noisy data.
\item Proving polynomial running time guarantees.
\end{compactenum}
To model noisy and imprecise data, such as sensor readings, we represent data as sets of intervals. Previous results \cite{PAC+SMT} only allowed for sets of assignments. We prove polynomial-time guarantees for implicitly learning any fragments of SMT with polynomial-time decision procedures, such as linear arithmetic. Since we are extending the PAC-Semantics framework, we are inheriting all of its useful features: e.g., these examples could be drawn an arbitrary distribution and are not required to be fully consistent with the learned example; nonetheless, we are able to give approximately correct answers with a tight guarantee. 

\subsection{Empirical Contributions}
\begin{compactenum}
\item Realising the first implementation of the PAC-Semantics framework.
\item Showing that implicit reasoning can be much faster and more noise-resistant than explicit methods for linear programs (LPs).
\end{compactenum}
We test our implementation on benchmark LP problems, where the task is to find the optimum objective value subject to the constraints of the problem. We compare the running times with an alternative approach of first creating an explicit model using IncaLP \cite{IncaLP} and then finding the optimum objective value based on the model. Our results show significantly lower running times using the implicit approach, and this advantage becomes more pronounced as the models grow. We also demonstrate that, in the presence of noisy data, implicit learning returns a feasible, approximate solution to the true linear program, whereas IncaLP often fails to find any model at all.

\section{Preliminaries}
We will now briefly review concepts from logic, satisfiability, PAC-Semantics and the formal setup used in our work.
\subsection{Logic}
\textit{Satisfiability} (SAT) is the problem of deciding whether there exists an assignment of truth values (i.e., model) to variables (propositional symbols) such that a propositional logical formula is true. SMT is a generalization to SAT for deciding satisfiability for fragments and extensions of first-order logic with equality, for which specialized decision procedures have been developed. Deciding satisfiability for these logics is done with respect to some decidable background theory  \cite{SMTbook}. In this work, we are especially interested in the background theories of quantifier-free linear real arithmetic (QFLRA). The following formal exposition is adapted from \cite{SMTbook}:

\paragraph{Syntax.} We assume a logical signature consisting of a set of predicate symbols, and a set of functions,  logical variables, and standard connectives ($ \wedge, \vee, \neg$). An atomic formula is one of the form: $b$ (a propositional symbol), $pred(t_1,...,t_k)$, $t_1 = t_2$, $\bot$ (false), $\top$ (true), where $t_1, ..., t_k$ are the terms of the signature, in the standard logical interpretation. A literal $l$ is an atomic formula or its negation $\neg l$. A ground expression is one where all the variables are replaced by constants and in general terms of the domain of discourse.

\paragraph{Semantics.} Formulas are given a truth value from the set $\{\bot, \top \}$ by means of first-order models.  A model $\pmb{\rho}$ can be seen simply as an element of $\Sigma ^n$, which is the universe of the model in $n$ dimensions. Throughout the paper we work with the signature of linear real arithmetic, with function symbols of the form $\{0,1,+,-,\leq,<,\geq,>,=,\neq\}$, interpreted in the usual way over the reals. 

\subsection{PAC-Semantics}

PAC-Semantics was introduced by Valiant \shortcite{Valiant2000} to capture the quality possessed by knowledge learned from independently drawn examples from some unknown distribution $D$. The output produced using this approach does not express validity in the traditional (Tarskian) sense. Instead, the notion of {\em validity} is then defined as follows:

\begin{definition}[$(1-\epsilon)$-validity \cite{Valiant2000}]
Given a joint distribution $D$ over $\Sigma^n$, we say that a Boolean function $f$ is $(1-\epsilon)$-valid if $\textrm{Pr}_{\pmb{\rho}\in D}[f(\pmb{\rho})=1]\geq1-\epsilon$. If $\epsilon=0$, we say that $f$ is perfectly valid.
\end{definition}

The reasoning problem of interest is deciding whether a query formula $\alpha$ is $(1-\epsilon)$-valid. Knowledge about the distribution $D$ comes from the set of examples $\pmb{\rho}$,  independently drawn from this distribution. Additional knowledge can come from a collection of axioms $\Delta$, known as the knowledge base (KB). 

In implicit learning, the query $\alpha$ is answered from examples directly, without creating an explicit model. This is done by means of entailment: we repeatedly ask whether $\Delta\wedge\pmb{\rho}^{(k)}\models\alpha$ for examples $\pmb{\rho}^{(k)}\in D$. If at least $(1-\epsilon)$ of the examples entail $\alpha$, we accept. The more examples we use, the more accurate our estimate is and the more confident we can be in it. The concepts of accuracy and confidence are captured by the hyper-parameters $\gamma, \delta \in (0,1)$, where $\gamma$ represents the accuracy of the examples used and $\delta$ captures the confidence of the example received. The number of examples needed will be determined from $\delta$ and $\gamma$; cf. Section \ref{sec:main} for a precise formulation.

Reasoning from complete examples is trivial: Hoeffding’s inequality \cite{hoeffding} guarantees that with high probability, the proportion of times that the query formula evaluates to ‘true’ is a good estimate of the degree of validity of that formula. To capture the more interesting problem of learning from incomplete examples, previous works used the notion of a masking process developed by Michael \shortcite{Loizos2010}. A masking process randomly replaces some of the values of variables in a formula with * before passing the examples to the learning algorithm. 

\begin{definition}[Masking process \cite{Loizos2010}] A {\em mask} is a function $M: \Sigma^n \rightarrow \{\Sigma \cup \{*\}\}^n$,  with the property that for any $\pmb{\rho} \in \Sigma^n$, $M(\pmb{\rho})$ is {\em consistent} with $\pmb{\rho}$, i.e., whenever $M(\pmb{\rho})_i \neq \ast $ then $M(\pmb{\rho})_i = \pmb{\rho}_i$. We refer to elements of the set $(\Sigma \cup \{ \ast \})^n$ as {\em partial assignments}. A {\em masking process} $\pmb{M}$ is a mask-valued random variable. We denote the distribution over partial examples obtained by applying the masking process as $\pmb{M}(D)$.
\end{definition}

\begin{example}
Assume a language with two real-valued variables $x$ and $y$. Given a full assignment $\pmb{\rho} : \{ x = 4, y = 5\}$, after applying a masking process on this assignment, we may obtain the partial assignment $\rho : \{x=4,y=*\}$.
\end{example}

In the above example, $x$ is not masked, and thus remains consistent with $\pmb{\rho}$, while $y$ gets masked. A partial model will be written using the regular font as $\rho$ vs the bold font for a full model as $\pmb{\rho}$. 

Work by \cite{PAC+SMT} showed that implicitly learning to reason can be done efficiently with partial assignments for standard fragments of arithmetic theories, which integrates a decision procedure to solve SMT formulas. They proposed a reduction from the learning to reason problem for a logic to any sound and complete solver for that fragment of logic. They showed that, as long as there is a sound and complete solver for a fragment of arithmetic, then reasoning under partial evaluation (denoted as $|_{\rho}$) can be obtained.

\begin{Theorem}[\cite{PAC+SMT}]\label{thm:complete-rc}
Let $\cal L$ be a logical language such that for any $\alpha\in\cal L$ and any partial assignment $\rho$ for $\cal L$, $\alpha|_\rho\in \cal L$ also.
Let A be a sound and complete procedure for deciding entailment for $\cal L$. Suppose $\Delta, \alpha \in \cal L$, and  $\rho$ is a partial assignment for $\cal L$. If $\Delta \models \alpha$, then $\Delta | _{\rho} \models \alpha | _ {\rho}$.
\end{Theorem}

They provide an algorithm (a less general version of DecidePAC as we discuss later) that enables implicit learning of fragments of SMT under partial assignments. We continue this line of work and present an extension of the assignments used in DecidePAC to sets of intervals. 

\section{Use Case Example}
\label{sec:fitness_example}

\begin{figure}[t]
\centering
\includegraphics[width=0.9\linewidth]{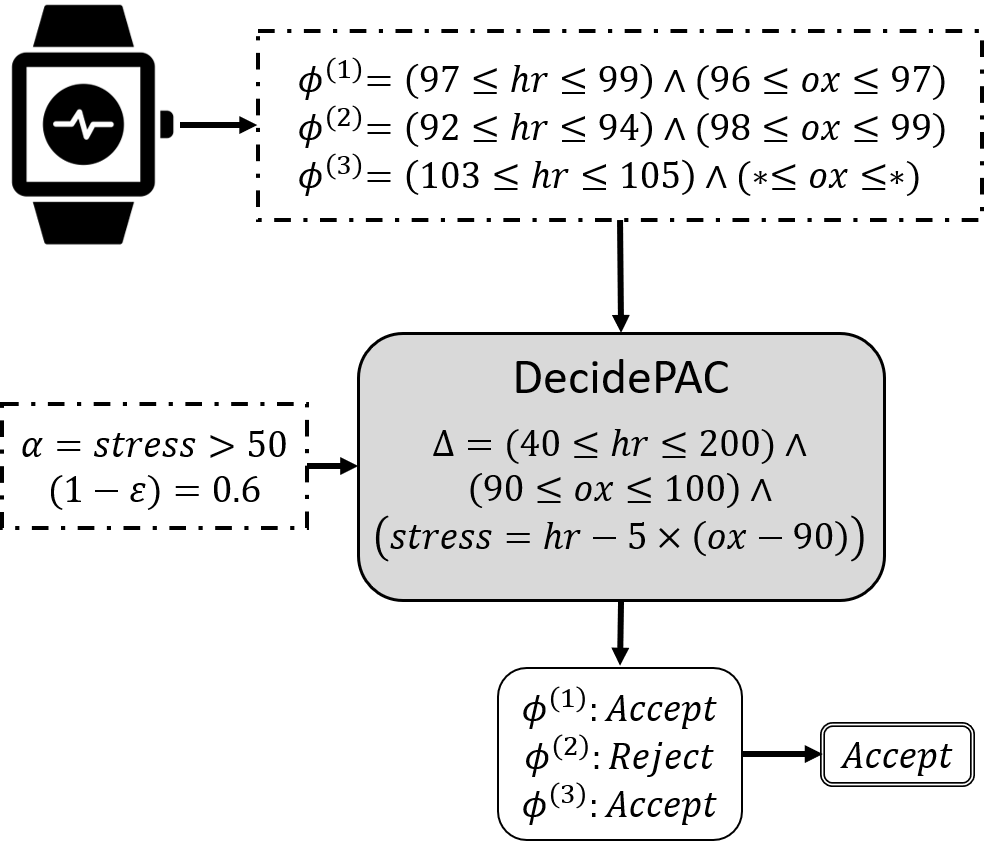}
\caption{High-level example of the framework}
\label{fig:pac_example_fitness}
\end{figure}

Let us demonstrate the functionality of the extended PAC framework with a small example: Consider a smart fitness watch, which monitors the heart rate and blood oxygen levels of the wearer. It calculates the wearer's stress level using a formula: $stress = hr - 5 \times (ox - 90)$, where \(hr\) is the heart rate in beats per minute and \(ox\) is the percentage of oxygen saturation in the blood.

As illustrated in Figure \ref{fig:pac_example_fitness}, this formula is encoded in the knowledge base of the system, $\Delta$, along with two chosen bounds for \(hr\) and \(ox\). The bound for \(ox\) is set between 90\% and 100\%, as anything lower is unnatural and a sign of hypoxemia \cite{hypoxemia}. Similarly, \(hr\) can be bounded by the values 40 and 200 \cite{heart_rate}. 

The watch alerts the user if the stress level exceeds 50, which is encoded with the query $\alpha: (stress > 50)$. Depending on the values of the heart rate and oxygen within the boundaries set in $\Delta$, the stress value might be below or above 50. Therefore, the explicit knowledge base alone cannot answer the query. However, the watch gets regular sensor readings in the form of intervals $\phi^{(k)}$, which it can use to answer the query.  Since the watch might not be tight enough on the wrist, no guarantees can be made about all the sensor readings being complete. In this example, some values for oxygen are missing. 

Using DecidePAC (cf. Algorithm \ref{algo:il}), the watch answers the query by means of entailment for each of the readings received and the knowledge base: $\Delta\wedge\phi^{(m)}\models\alpha$.
It receives the readings $\phi^{(1)}, \phi^{(2)}, \phi^{(3)}$, out of which the first and third yield entailment, whereas the second does not (because $\{hr=92,ox=99\}$ leads to $stress=47 <50$). The third set of intervals, $\phi^{(3)}$, triggers validity despite missing data, as the stress level is above 50 for any possible values of \(ox\). Our validity parameter $\epsilon = 0.4$ allows the algorithm to return acceptance with $(1-\epsilon)$ validity of the query after seeing two out of three positive readings.

\section{Extending the Learning Model}
\label{sec:main}

\begin{algorithm}[t]
\KwIn{Procedure A, query $ \alpha$, validity $\epsilon\in (0,1) $, list of partial intervals $\{\phi^{(1)}, ... , \phi^{(m)} \}$, knowledge base $\Delta$}
\KwOut{Accept/Reject}
\BlankLine
\Begin{
$B \leftarrow \lfloor \epsilon \times m \rfloor , FAILED \leftarrow 0$.\\

\ForEach{$k$ in $1..m$} {
\If{$A(\alpha,\phi^{(k)}, \Delta)$ returns UNSAT}{
Increment $FAILED$.\\
\lIf{$FAILED > B$}{\bf return Reject}}
}
\bf return  Accept \\}
\caption{DecidePAC}
\label{algo:il}
\end{algorithm}

In this section, we extend the framework to deal with intervals as opposed to assignments and prove that polynomial-time guarantees are preserved. We refer to a masking process in the context of intervals $\phi$ as a blurring process, as introduced below\footnote{We use superscript to represent the index of the assignment, and subscript to represent the index of the variable in each assignment.}:

\begin{definition}[Blurring process]
Given a full assignment  $\pmb{\rho}^{(k)} = \{ \pmb{\rho_1}, \ldots ,\pmb{\rho_n } \}$, a \textit{blurring} function is defined as $B: \Sigma^n\rightarrow\{\Sigma \cup \{-\infty, +\infty\}\}^{2n}$, which produces a set of intervals $\phi^{(k)}$ consistent with the assignment $\pmb{\rho}^{(k)}$, i.e., with the property that for each $\pmb{\rho}_i$, $B(\pmb{\rho})_{2i-1}\leq\pmb{\rho}_i\leq B(\pmb{\rho})_{2i}$, where $B(\pmb{\rho})_{2i-1}$ is a random value from $( - \infty, \pmb{\rho}_i]$ marking the lower bound of the interval and $B(\pmb{\rho})_{2i}$ is a random value from $[ \pmb{\rho}_i, \infty)$ marking the upper bound. We refer to elements of the resulting set as \textit{partial intervals}, where a full assignment is bound by the lower and upper bound. A \textit{blurring process} $\textbf{B}$ is a blur-valued random variable (i.e. a random function). We denote the distribution over intervals obtained by applying the blurring process as $\textbf{B}(D)$.
\end{definition}

The blurring function essentially offers a generalization of the masking process in order to model uncertainty about complete assignments.
For each interval, we draw one lower and one upper bound, hence the endpoints of the $n^{th}$ interval occupy indices $2n-1$ and $2n$. Using the blurring process, we allow observations to be not only (i) unknown (masked) but also (ii) uncertain, and this degree of uncertainty is given by the width of the interval in which the real value lies.

\begin{example}
Given a full assignment $\pmb{\rho} : \{ x = 4, y = 5\}$, after applying a blurring process on this assignment, we may obtain the intervals set $\phi : \{( 1 \leq x) ,(x \leq 6),(-\infty \leq y), (y \leq \infty) \}$.
\end{example}

In the above example, the value of $x$ has increased in uncertainty, in the sense that we have a small range of possible values. The value of $y$ has been extended on both sides to $\infty$ and denotes complete uncertainty (equivalent to the effect of a masking process). In other words, the relaxation of a fixed value/assignment to the interval directly corresponds to the uncertainty about the value. 

The output of a blurring process is a list of partial intervals. In order to integrate these observations in a decision procedure, i.e., entailment, they require an additional transformation into logical formulas:

\begin{definition}[Grounding  $\downarrow$] 
Given a partial interval $\phi^{(k)} = \{ \phi_1, ...,\phi_{2n}\}$, we define $\phi^{(k)} \downarrow$ as the ground formula corresponding to the set of intervals and represented as a set of constraints, i.e., $\phi^{(k)} \downarrow = \phi_1 \wedge ... \wedge \phi_{2n}$.
\end{definition}

In the evaluation procedure, we define a formula to be witnessed to evaluate to \textit{true} given partial intervals as follows:
 
\begin{definition}[Witnessed formulas]
A formula $\varphi$ is witnessed true under partial intervals $\phi$ if $(\phi\downarrow)\models\varphi$ , i.e. $\varphi$ is true under every assignment possible under $\phi$.
\end{definition}

In \cite{PAC+SMT}, it was shown that, if $\Delta \models \alpha$, then $\Delta |_{\rho} \models \alpha |_{\rho}$, where $\rho$ is a partial assignment. The key difference in our work is that instead of partial assignments, we use partial intervals. However, since partial intervals are well defined formulas of QFLRA, this property directly extends to our case and the proof is analogous to \cite{PAC+SMT}. 

\begin{proposition}
For the language of QFLRA, let $A$ be a sound and complete decision procedure. If $\Delta\models\alpha$, then $\Delta\vert_{\phi}\models\alpha\vert_{\phi}$. Equivalently, if $A(\Delta,\alpha)$, then $A(\Delta\vert_{\phi},\alpha\vert_{\phi})$ iff $A(\Delta\wedge\phi\downarrow,\alpha)$.
\end{proposition}

We can now prove that implicit learning remains possible with our extended framework and that existing polynomial-time guarantees from the assignment setting are preserved:

\begin{Theorem} 
[Implicit learning]\label{thm:il}
Let $\Delta$ be a conjunction of constraints representing the knowledge base and $\alpha$ an input query. We draw at random $m = \frac{1}{2\gamma^2}ln\frac{1}{\delta}$ sets of intervals $\{\phi^{(1)}, \phi^{(2)}, ..., \phi^{(m)}\}$ from $\pmb{B}(D)$ for the distribution $D$ and a blurring process $\pmb{B}$. Suppose that we have a decision procedure $A$. Then with probability $1-\delta$:
\begin{compactitem}
  \item If $(\Delta \Rightarrow \alpha)$ is not $(1- \epsilon - \gamma)$ - valid with respect to the distribution $D$, the DecidePAC algorithm returns Reject; and
    \item If there exists some KB $I$ such that $\Delta \wedge I \models \alpha$ and $I$ is witnessed true with probability at least $(1 - \epsilon + \gamma)$ on $\pmb{B}(D)$, then DecidePAC returns Accept.
\end{compactitem}
Moreover, if $A$ runs in polynomial-time (in the number of variables, size of the query, and size of the knowledge base), so does DecidePAC.
\end{Theorem}

\begin{proof}
Consider a sound and complete decision procedure $A$ for the language domain aforementioned and the reasoning problem of deciding $\Delta \models \alpha$.   By definition of soundness and completeness, $\Delta \models \alpha$ if and only if $A(\Delta \wedge \neg \alpha)$ = UNSAT. Suppose we receive observations about the world as sets of blurred intervals $\phi$ and we wish to decide entailment of the aforementioned problem with respect to these blurred observations, hence calculate  $A(\Delta | _{\phi} \wedge \neg \alpha | _{ \phi})$ = UNSAT. This holds iff $A(\Delta \wedge \phi \downarrow \wedge \neg \alpha)$ = UNSAT, iff $A(\Delta \wedge I \wedge \phi \downarrow \wedge \neg \alpha)$ = UNSAT for any KB $I$ that is witnessed true under $\phi$. Now the argument becomes analogous to Juba \shortcite{juba2013}: if $\Delta\Rightarrow\alpha$ is unsatisfied on a point drawn from $D$, it is not entailed by the blurred example from $\textbf{B}(D)$ either, so FAILED increments when such points are drawn, and does not increment when a suitable $I$ is witnessed true. By Hoeffding's inequality, the returned value satisfies the given conditions with probability $1-\delta$ for $m$ examples. The decision procedure will return UNSAT in polynomial time $T(n)$ depending on the size of the knowledge base and query. Every iteration costs the time for checking feasibility which is bounded by the time complexity of the decision procedure used for deciding satisfiability. The total number of iterations is $m=\frac{1}{2\gamma^2} \log\frac{1}{\delta}$, corresponding to the number of samples drawn, which gives us the total time bound of $O(T(n)\cdot \frac{1}{\gamma^2} \log\frac{1}{\delta})$.
\qed
\end{proof}

With this theorem, we have extended the previous result to deal with the more complex setting of threshold uncertainty, which is a much more realistic learning model. In QFLRA, the decision procedure is polynomial-time \cite{SMTbook} and blurred examples have the natural interpretation of being imprecise measurements. However, Theorem \ref{thm:il} is in principle applicable to any decidable fragment of SMT. Interpreting blurred examples for other domains (e.g., as noisy observations in the form of disjunctions of partial models) is certainly interesting, but will be left as a direction for the future. 

Like the previous PAC-Semantics results, implicit learning is geared for answering queries, and has been primarily a theoretical framework. In contrast, conventional methods, although not always offering learnability guarantees, provide an explicit model, making it harder to compare with the above framework and study the trade-offs. The next section shows how this is now possible. 

\section{Applying PAC-Semantics to Optimisation}

\begin{algorithm}[ht!]
\KwIn{Procedure A, preference function $f$, validity $\epsilon \in (0,1)$, accuracy $a \in \mathbb{Z}^+$, list of intervals $\phi = \{\phi^{(1)}, \phi^{(2)}, ... , \phi^{(m)} \}$, goal $\in$ \{"max", "min"\}}
\KwOut{estimated optimal value w.r.t. $f$}
\Begin{
\lIf{goal = "min"}{
$f \leftarrow -f$}
\eIf{DecidePAC(A, $0 \geq f, \epsilon,\phi$) accepts}{
\eIf{DecidePAC(A, $-1 \geq f, \epsilon,\phi$) rejects}{
$l \leftarrow -1, u \leftarrow 0$}{
$b \leftarrow -2$ \\
\While{DecidePAC(A, $b \geq f, \epsilon,\phi$) accepts}{
$b \leftarrow b \times 2$}
$l \leftarrow b, u \leftarrow b / 2$}}{
\eIf{DecidePAC(A, $1 \geq f, \epsilon,\phi$) accepts}{
$l \leftarrow 0, u \leftarrow 1$}{
$b \leftarrow 2$ \\
\While{DecidePAC(A, $b \geq f, \epsilon,\phi$) rejects}{
$b \leftarrow b \times 2$}
$l \leftarrow b/2, u \leftarrow b$}}{
}
\For{$a$ iterations}{
\eIf{DecidePAC(A, $(l+u)/2 \geq f, \epsilon,\phi$) accepts}{
$u \leftarrow (l+u)/2$}{
$l \leftarrow (l+u)/2$}}
\leIf{goal = "min"}{
\Return $-l$} 
{\Return $l$ }
} 
\caption{OptimisePAC}
\label{algo:binary}
\end{algorithm}

The goal in optimisation problems is to find an optimal objective value given constraints. We propose Algorithm \ref{algo:binary}, OptimisePAC, which solves optimisation problems using DecidePAC. This allows us to compare our framework to other more traditional methods, since optimisation problems are widely studied. In particular, we will use OptimisePAC in our empirical analysis for the case of LPs. The correctness and running time are as follows:

\begin{Theorem}
Let $\Delta$ be a conjunction of constraints representing the knowledge base and as input preference function $f$. We draw at random $m = O(\frac{1}{ \gamma^2}\log \frac{1}{\delta})$ partial intervals $\phi^{(1)},..., \phi^{(m)}$ from $\pmb{B}(D)$ for a distribution $D$ and a blurring process $\pmb{B}$.
Suppose that we have a decision procedure A running in time $T(n)$. Then, the OptimisePAC algorithm will return  $a$  significant bits of a value $v^*$ that is attainable on $I\wedge\Delta$
 for all KBs $I$ that are witnessed with probability $1-\epsilon+\gamma$, and such that for the value $u^*$ obtained by incrementing the $a^{th}$ bit, $\Delta \Rightarrow (f\leq u^*)$ is $(1-\epsilon-\gamma)$-valid (resp., $f\geq u^*$ with the $a^{th}$ bit decreased if minimising)
in time $O(T(n) \cdot m \cdot a)$.
\end{Theorem}

\begin{proof}
OptimisePAC first finds the approximate bounds of the optimal value by doubling the bound each iteration and running DecidePAC. If a lower bound $b/2$ and an upper bound $b$ are found, it runs binary search to find the optimal value to the desired precision. If the decision procedure runs in time $T(n)$, the total running time stated in the theorem follows.

To prove correctness, we will use a theorem due to Talagrand \shortcite{talagrand1994} (Theorem 4.9 of \cite{NNLearningbook}): 

\begin{Theorem}
[\cite[Theorem 4.9]{NNLearningbook}] There are positive constants $c_1,c_2, c_3$ such that the following holds. Suppose that $F$ is a set of functions defined on the domain $X$ and that $F$ has a finite VC dimension $d$. Let $ \gamma \in (0,1)$ and $m \in \mathbb{Z}^+$. Then the probability that the empirical mean of any $f\in F$ on $m$ examples differs from its expectation by more than $\gamma$ is at most $ c_1 c_2^d e ^{-c_3 \gamma^2 m}$.
\end{Theorem}

Thus, for $m \geq \frac{c_3}{\gamma^2}(d\ln c_2 + \ln \frac{c_1}{\delta})$, the bound is at most $\delta$.

Recall, the VC dimension is the size of the largest set of points that can be given all labels by a class (``shattered''). Consider a fixed class of Boolean functions on the blurred samples, which is parameterized by the objective value bounds $b$. This function outputs the value 1 whenever $\Delta \wedge \phi \wedge (f(x) \leq b)$ returns UNSAT, and 0 otherwise. We will show that this class has VC-dimension at most $1$.

We will show that for any two blurred  examples $\phi_1$ and $\phi_2$, it is not possible to get all the labellings $\{(1,1), (1,0), (0,1), (0,0)\}$ by varying $b$. 
Suppose there is some $b^*$ for which $\phi_1$ gives the label 1 and $\phi_2$ gives 0, meaning that for $\phi_1$ the bound $f \leq b^*$ does not hold and for $\phi_2$ it does.
Since $f \leq b^*$ holds for $\phi_2$, then for any $b > b^*$, the decision procedure will return 0 for $\phi_2$. On the other hand, the bound $f \leq b^*$ will not hold for $\phi_1$ for all values $b < b^*$.
Thus, in either direction, one of the labels for one of the two remains the same. So, it is not possible to get all possible labellings of $\phi_1$ and $\phi_2$ and so the VC-dimension is $\leq 1$.

Therefore, by Talagrand's bound, given $m$ examples, with probability $1-\delta$, DecidePAC answers correctly for all queries made by OptimisePAC. In particular, we note that the algorithm maintains the invariant that $l$ is the largest value for which $l\geq f$ was rejected by DecidePAC. Since it was not accepted, we see that for any $I$ that is witnessed with probability $\geq 1-\epsilon+\gamma$, there must exist some $x$ satisfying $I\wedge\Delta$ with $f(x)>l$ (resp., $f(x)<l$ if minimising). Since DecidePAC does not reject with $u$, $\Delta\Rightarrow f(x)\leq u$ is $(1-\epsilon-\gamma)$-valid, where $u$ and $l$ only differ by the value of the $a^{th}$ most significant bit. Thus $l$ is as needed.
\qed
\end{proof}

\section{Empirical Analysis}
\begin{figure*}[ht]
  \centering
  \begin{subfigure}[b]{1\textwidth}
  \includegraphics[width=1\textwidth]{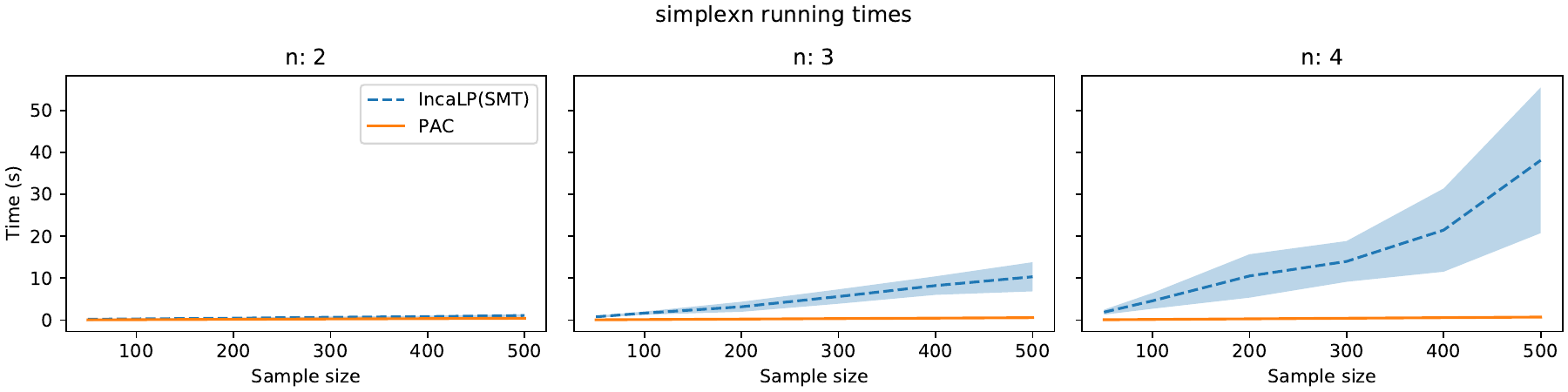}
  \label{fig:simplexn_runtimes}
  \end{subfigure}
  \begin{subfigure}[b]{1\textwidth}
  \includegraphics[width=1\textwidth]{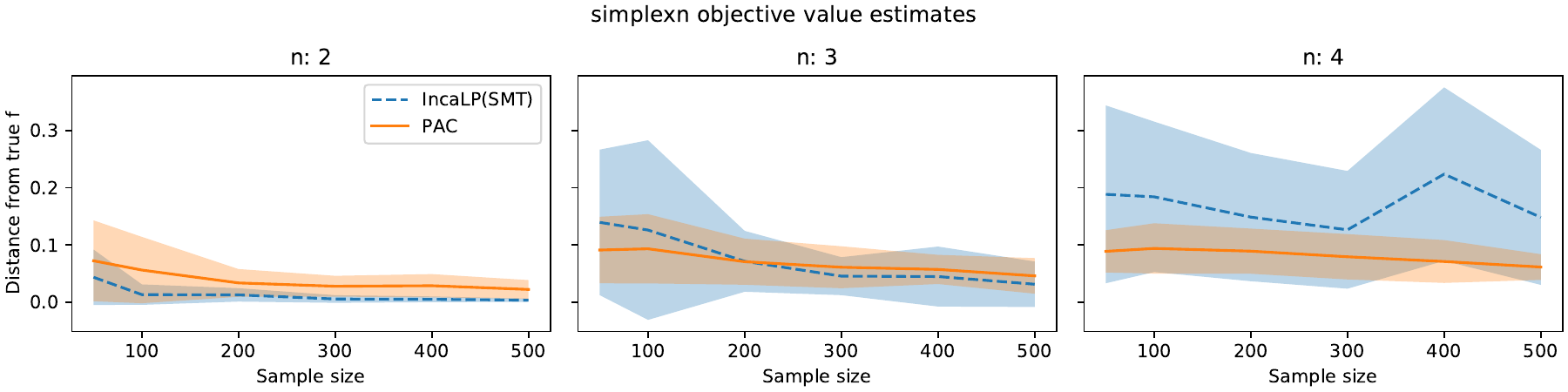}
  \label{fig:simplexn_fs}
  \end{subfigure}
  \caption{Comparing running times and objective value estimates for \textit{simplexn}. Here, \textit{n} is the number of dimensions, \textit{distance from true f} is the absolute difference between the true and the estimated objective value.}
  \label{fig:simplexn}
\end{figure*}

Implicit learning in PAC-Semantics has been described rigorously on a theoretical level. However, to the best of our knowledge, the framework has never been analysed in practice. In this section, we present the first implementation of the PAC-Semantics framework.\footnote{The code can be found at\\ \texttt{\url{https://github.com/APRader/pac-smt-arithmetic}}} We will use it to empirically study the differences between an implicit and explicit approach for LP problems. LP is an optimisation technique for a linear objective function. The problem consists of an optimisation function $f(\vec{x})$, where $\vec{x} \in \mathbb{R}^n$, and feasibility region expressed as the set of constraints $A \cdot \vec{x} \leq b$.  We chose LPs as the domain, as they can be represented in QFLRA, and are thus polynomial-time. It also allows us to directly evaluate our framework using existing benchmarks. We will analyse the following four points: running time, accuracy of objective value estimates, noise and outlier resistance.

\subsection{Implementation Setup}
\begin{figure*}[ht]
\centering
\includegraphics[width=\textwidth]{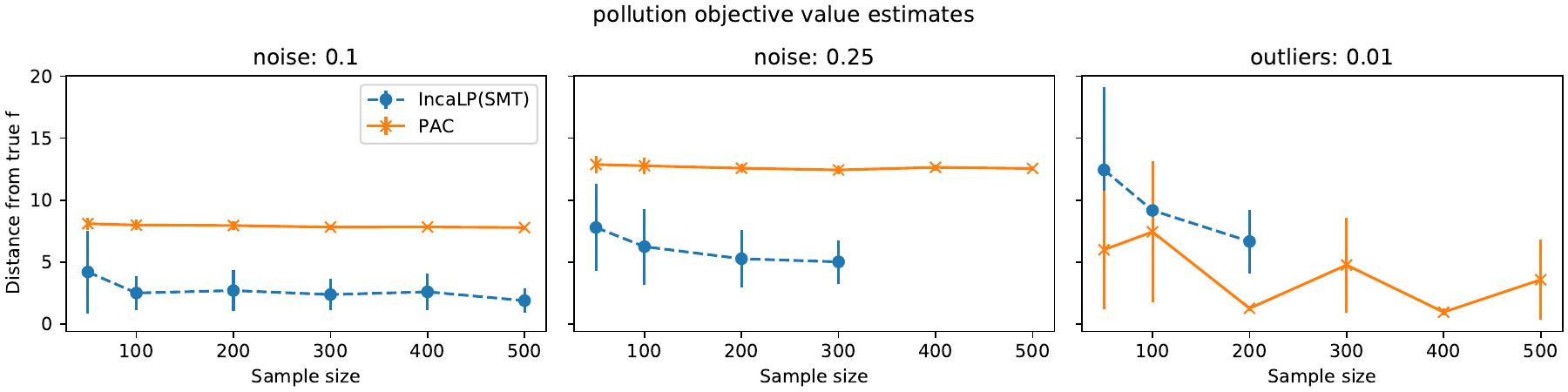}
\caption{Objective values for \textit{pollution}. True minimum: 32.15. IncaLP failed to find a model 30, 71 and 82 out of 120 times respectively.}
\label{fig:pollution}
\end{figure*}

The implementation is written in Python and uses the Z3 Theorem Solver \cite{Z3} for SMT queries. The explicit solver we compare it to is IncaLP(SMT), an algorithm that learns linear arithmetic SMT formulas from data \cite{IncaLP}. In this paper, we refer to IncaLP(SMT) simply as IncaLP. 

We use the following benchmark LP problems for our analysis: \textbf{simplexn}, \textbf{cuben}, \textbf{pollution} and \textbf{police} \cite{textbook_examples}. All these problems consist of a set of hard constraints, which define the feasible region or the boundaries of the shape and a set of soft constraints in the form of a linear objective function, which can be either maximised or minimised. The goal is to find the optimum objective value within the hard constraints. For example, the feasible region in \textit{simplexn} is an intersection of $\frac{1}{2}n(n-1)$ uniform triangular prisms and the objective function for \textit{pollution} represents the amount of pollution and has to be minimised. 

The implicit PAC approach and the explicit IncaLP approach differ in how they reach this goal: in the PAC model, we directly compute the optimal objective value from positive examples using OptimisePAC. In contrast, the IncaLP approach first creates an SMT model from positive and negative examples. Using standard MaxSMT techniques, we can then find the optimal objective value within the model. 

We ran tests on all four problems. For each test, we had 20 independent runs on increasing sample sizes from 50 to 500 and increasing dimensions from 2 to 4, if applicable. We used 50\% positive and 50\% negative samples. A timeout of 30 minutes was set for each run. To ensure reproducibility, we ran each test with the seed 1613 (a numerical encoding of ``PAC"). The hardware we used was an Intel Core i7-6700 3.40GHz, with 16 GB of RAM running Ubuntu 20.04.

As for parameter settings, we chose the SelectDT heuristic for the IncaLP approach, as it is the most appropriate according to their paper. We also ran our experiments using the same initial configurations as their released code. We set the accuracy for OptimisePAC to 60, which is the number of times our intervals are divided by 2. The reason being that we can match the accuracy of double precision floating-point values, which is about $2^{-53}$. 
\subsection{Results}

On the theoretical front, one of the advantages of implicit learning is argued to be efficiency, since you can skip the step of creating an explicit model. As shown in Figure \ref{fig:simplexn}, this effect is significant in practice for linear programs. PAC is able to get similarly good objective value estimates at significantly lower running times for \textit{simplexn}. The larger the sample size and the higher the dimensionality, the bigger the gap between running times. (For all of our graphs, the sample standard deviation is shown.)

With the extension we introduced in this paper, PAC can now handle noisy data using intervals. If we add Gaussian noise with a standard deviation $\sigma$ to each data point, we can represent it using intervals of width $4 \log{d} \cdot \sigma$, where \(d\) is the dimensionality. This interval covers about 95\% of the density, which is why we set a validity of 95\% for DecidePAC. The noise and intervals were capped at the domain boundaries. We adjusted the noise to the dimensionality, meaning that for a noise value of \(n\), the std of the Gaussian $\sigma=\frac{n}{\sqrt{d}}$. We also tested robustness against outliers. For our purposes, an outlier is a point within the domain with a wrong label.

Figure \ref{fig:pollution} shows how PAC compares to IncaLP. In cases where IncaLP does find a model, it gives an estimate that is closer to the true objective value in noisy cases. However, it failed to find a model 25\%, 59\% and 68\% of the time for a noise of 0.1, 0.25 and outliers of 0.01, respectively. Moreover, the PAC estimate is pessimistic on purpose. This results in values farther away from the optimum but ensures that they are reachable. PAC always finds feasible objective values, while IncaLP undershoots the minimum for \textit{pollution }almost always.

It is also worth noting these results do not mean that IncaLP, on the whole, is inferior to our PAC implementation. IncaLP creates a model, while PAC answers queries implicitly. For some problems, such as objective value optimisation, it is possible to skip the model making process. As we have shown, doing that comes with advantages in speed and noise resistance. However, in some contexts, having an explicit model is desirable, in which case the implicit learning paradigm might be harder to work with. Put simply, OptimisePAC is not a replacement for IncaLP.

\section{Related Work}

We identify two threads of related research. On the theoretical side, 
the problem of dealing with missing information in learning settings was recognized early on in the literature. For example, to analyze how incomplete data affects  predictions, \cite{Schuurmans94learningdefault} look at various ways of altering the input data, according to a product distribution or via masking. In \cite{KearnsShapire},  a framework is proposed that uses incomplete observations to predict a hypothesis function and never output a \textit{``don't know"} answer. These and related approaches, however, focus on a discrete model  and do not cover continuous-valued domains. There is, of course, also a large body of work on techniques like imputation, but these are far afield from our goals.  

Conceptually, it is perhaps also interesting to contrast this line of work with \textit{inductive logic programming} (ILP) \cite{Muggleton1994}: the latter searches for a hypothesis $H$ that is consistent with the examples by appealing to entailment. Like IncaLP, it does not, however, seek to analyze the degree to which the resulting formulas capture an unknown, ground-truth process that produced the examples. For more discussions on how implicit learning via the PAC-Semantics differs from such hypothesis generation methodologies, including probabilistic structure learning schemes, see \cite{juba2013,bellejuba2019}.  
 
On the empirical side, many  methods for inducing linear constraints from data use greedy or approximate learning schemes, e.g., \cite{large}. To overcome the problems of overfitting, local optima or noise-robustness, approaches like syntax-guided learning \cite{aluretal2013} have been proposed, in which the learning problem is reduced to an optimisation problem. See \cite{IncaLP} for a comprehensive discussion. Similar to IncaLP, work by \cite{Pawlak2017} proposes an approach of approximating the set of constraints from feasible and infeasible examples, which focuses on mixed integer linear programs, but without any theoretical guarantees.

\section{Conclusion}

In this work, we proposed a general framework for learning implicitly in linear arithmetic from noisy examples. By then considering a novel optimisation variant, we were able to empirically compare and outperform an explicit approach in terms of running time and resistance to noise and outliers for LP problems. A natural direction for the future is to consider whether this implementation strategy extends to other classes of formulas in first-order logic and/or SMT. In particular, it is worth conducting empirical investigations into polynomial-time fragments of SMT other than LPs.

\section*{Acknowledgements}
We would like to thank our reviewers for their helpful suggestions. Alexander Rader conducted this work while at the University of Edinburgh. Ionela Georgiana Mocanu was supported by the Engineering and Physical Sciences Research Council (EPSRC) Centre for Doctoral Training in Pervasive Parallelism (grant EP/L01503X/1) at the University of Edinburgh, School of Informatics. Vaishak Belle was supported  by a Royal Society University Research Fellowship. Brendan Juba was supported by NSF/Amazon award IIS-1939677.

\bibliographystyle{named}
\bibliography{ijcai21}

\appendix
\section{Experimental setup}
All experiments can be run from the command line using the file \texttt{incalp\_comparison.py}. It uses the following structure:
\begin{lstlisting}[language=bash]
python3 incalp_comparison.py problem [optional commands]
\end{lstlisting}

where \texttt{problem} is either \texttt{simplexn}, \texttt{cuben}, \texttt{pollution} or \texttt{police}.
\subsection{Optional commands}
\begin{itemize}
\item \texttt{-s, --seed}: The seed for the random number generator.
\item \texttt{-n, --noise}: Add Gaussian noise to samples with given standard deviation, normalised by number of dimensions.
\item \texttt{-o, --outliers}: Ratio of outliers, between 0 and 1.
\item \texttt{-t, --timeout}: Timeout for IncaLP in seconds. Only works for Unix-based systems.
\item \texttt{-v, --verbose}: Turn on verbose mode.
\end{itemize}
\subsection{Additional information}
All the necessary functions from IncaLP are in the folder \texttt{incalp}. The IncaLP source code was taken from \url{https://github.com/samuelkolb/incal/releases}. In addition, the following Python packages need to be installed: PySMT, matplotlib, numpy, scikit-learn, z3-solver.

As for parameter settings, we chose the SelectDT heuristic for the IncaLP approach, as it is the most appropriate according to their paper. We also ran our experiments using the same initial configurations as their released code. We set the accuracy for OptimisePAC to 60, which is the number of times our intervals are divided by 2. The reason being that we can match the accuracy of double precision floating-point values, which is about $2^{-53}$. 
\subsection{Reproducing our results}
For every experiment, we set the seed to 1613, a numerical encoding of the word "PAC". The sample sizes are always [50, 100, 200, 300, 400, 500] and the number of runs is 20. To be as accurate as possible, we ran all experiments that we put in the paper without a timeout. We used the following commands:

\begin{lstlisting}[language=bash]
python3 incalp_comparison.py simplexn --seed 1613
python3 incalp_comparison.py pollution --seed 1613 --noise 0.1
python3 incalp_comparison.py pollution --seed 1613 --noise 0.25
python3 incalp_comparison.py pollution --seed 1613 --outliers 0.01
\end{lstlisting}

Using the same command will result in the same samples and objective value estimates, since we set the seed. However, your running times might differ from ours, since you might be using different hardware and other processes on the machine can also affect the running time.

To be able to test a wide range of examples, we set a timeout of 1800 seconds (30 minutes) for the rest of the experiments. This is because the running times for IncaLP vary widely. For example, one run in police with 300 samples and 0.01 outliers took less than 12 minutes before finding a model, while the next ran for over 5 hours and did not find a model anyway. The commands for these test are below:

\begin{lstlisting}[language=bash]
python3 incalp_comparison.py simplexn --seed 1613 --timeout 1800 --noise 0.1
python3 incalp_comparison.py simplexn --seed 1613 --timeout 1800 --noise 0.25
python3 incalp_comparison.py simplexn --seed 1613 --timeout 1800 --outliers 0.01 
python3 incalp_comparison.py cuben --seed 1613 --timeout 1800
python3 incalp_comparison.py cuben --seed 1613 --timeout 1800 --noise 0.1
python3 incalp_comparison.py cuben --seed 1613 --timeout 1800 --noise 0.25
python3 incalp_comparison.py cuben --seed 1613 --timeout 1800 --outliers 0.01
python3 incalp_comparison.py pollution --seed 1613 --timeout 1800
python3 incalp_comparison.py police --seed 1613 --timeout 1800
python3 incalp_comparison.py police --seed 1613 --timeout 1800 --noise 0.1
python3 incalp_comparison.py police --seed 1613 --timeout 1800 --noise 0.25
python3 incalp_comparison.py police --seed 1613 --timeout 1800 --outliers 0.01
\end{lstlisting}

Note that the timeout means that results cease to be exactly reproducible. A run might time out on a slower machine, but finish running on a faster one. Nevertheless, the results will, on average, be very similar.

\section{Results}
Here we show all the graphs that result from the commands listed above. Note that the following hold:
\begin{itemize}
\item The true minimum objective value for pollution is about 32.15 and for police it is 3.37.
\item The objective functions for simplexn and cuben are randomly generated and vary for each run. They are a linear combination of all variables with coefficients and a constant, each varying from -1 to 1. Thus, the true objective value $f$ is bounded by $-(d+1) \leq f \leq d+1$, where $d$ is the number of dimensions. The goal is to maximise that value.
\item For a noise parameter $n$, the standard deviation $\sigma$ of the Gaussian noise is adjusted by the dimensionality:  $\sigma=\frac{n}{\sqrt{d}}$.
\item Simplexn and cuben have dimensions ranging from 2 to 4. Pollution has 6 dimensions and police has 5. 
\end{itemize}
Detailed results for each experiment can be found in their respective logs, which are located in the \texttt{output} folder.

\begin{figure*}[ht]
  \centering
  \begin{subfigure}[b]{1\textwidth}
  \includegraphics[width=1\textwidth]{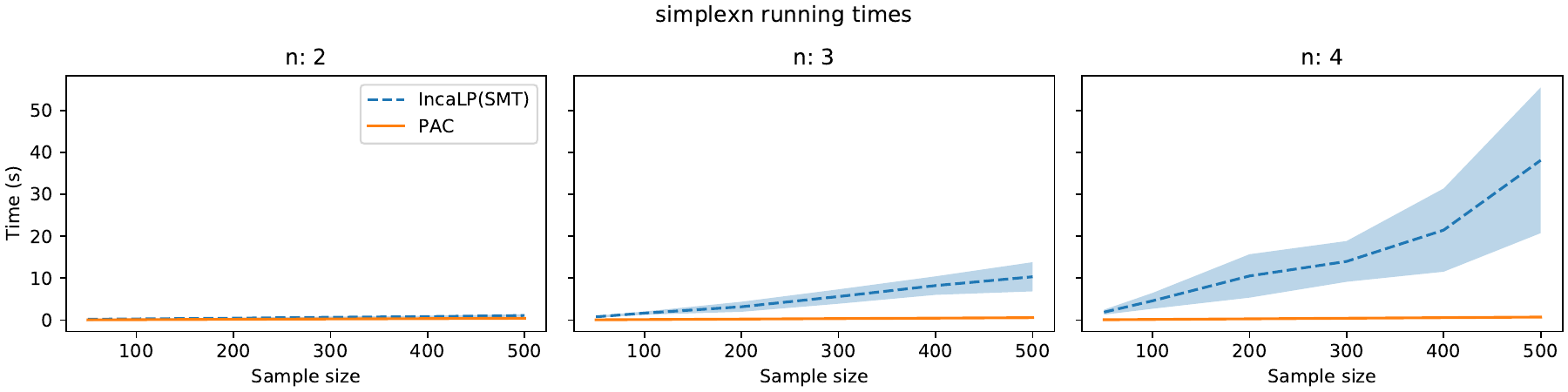}
  \end{subfigure}
  \begin{subfigure}[b]{1\textwidth}
  \includegraphics[width=1\textwidth]{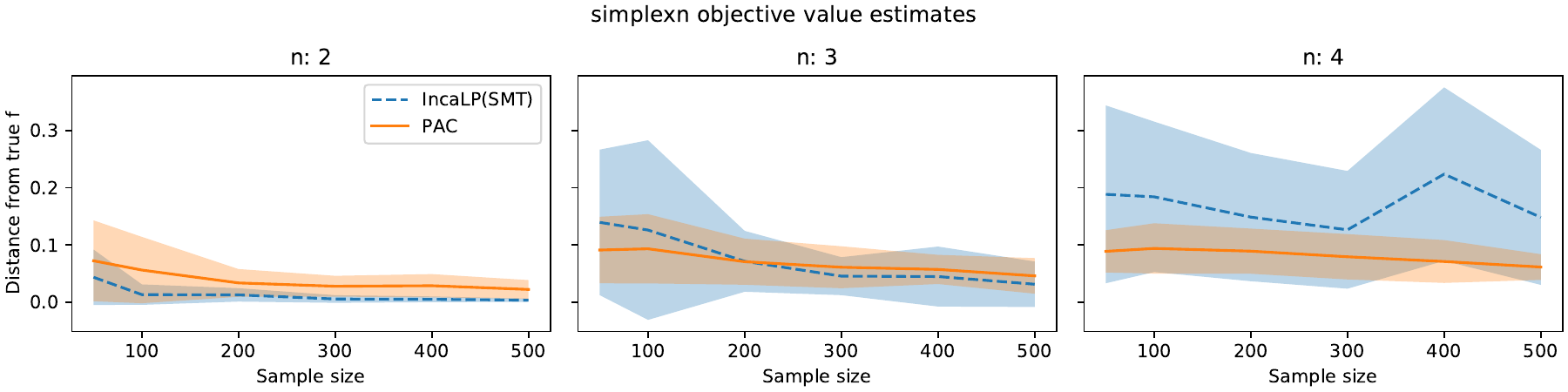}
  \end{subfigure}
  \caption{IncaLP always found a model.}
\end{figure*}

\begin{figure*}[ht]
  \centering
  \begin{subfigure}[b]{1\textwidth}
  \includegraphics[width=1\textwidth]{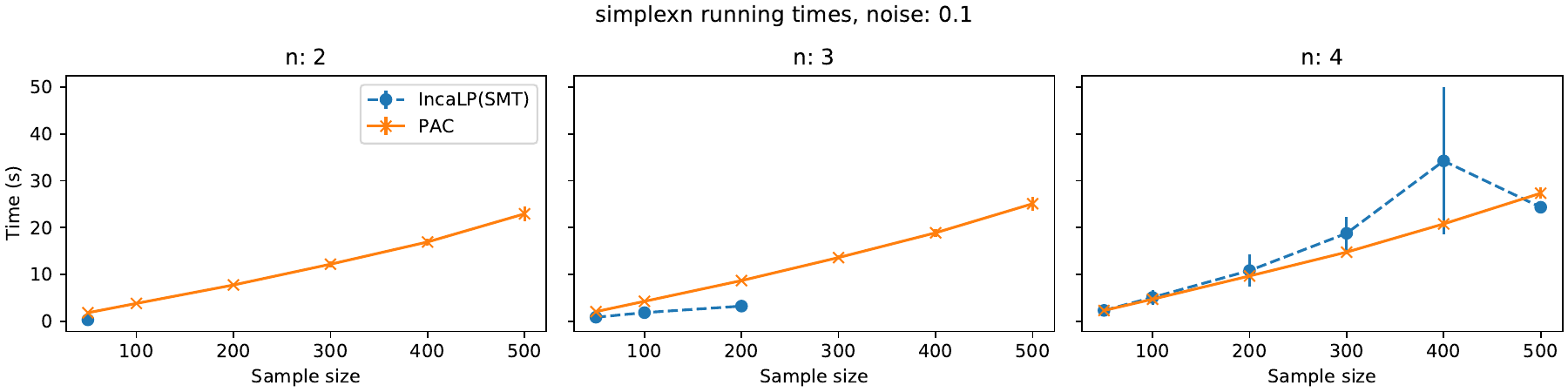}
  \end{subfigure}
  \begin{subfigure}[b]{1\textwidth}
  \includegraphics[width=1\textwidth]{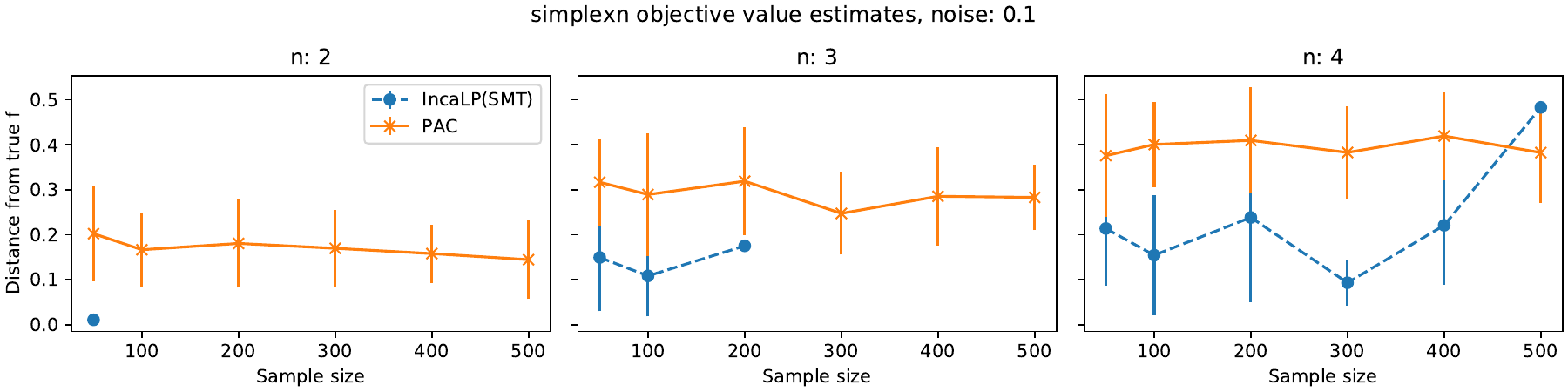}
  \end{subfigure}
  \caption{IncaLP failed to find a model 73\% of the time.}
\end{figure*}

\begin{figure*}[ht]
  \centering
  \begin{subfigure}[b]{1\textwidth}
  \includegraphics[width=1\textwidth]{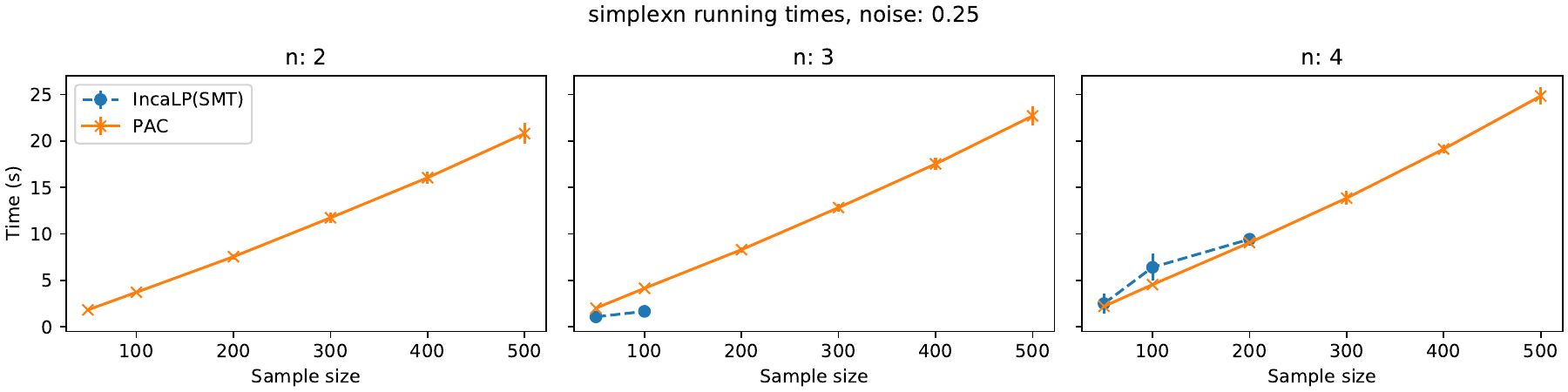}
  \end{subfigure}
  \begin{subfigure}[b]{1\textwidth}
  \includegraphics[width=1\textwidth]{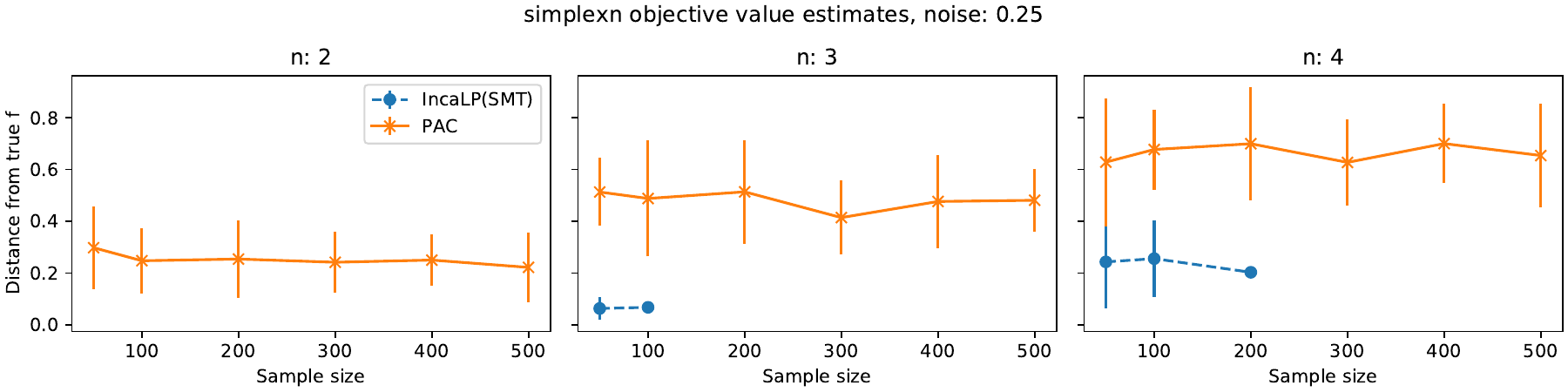}
  \end{subfigure}
  \caption{IncaLP failed to find a model 89\% of the time.}
\end{figure*}

\begin{figure*}[ht]
  \centering
  \begin{subfigure}[b]{1\textwidth}
  \includegraphics[width=1\textwidth]{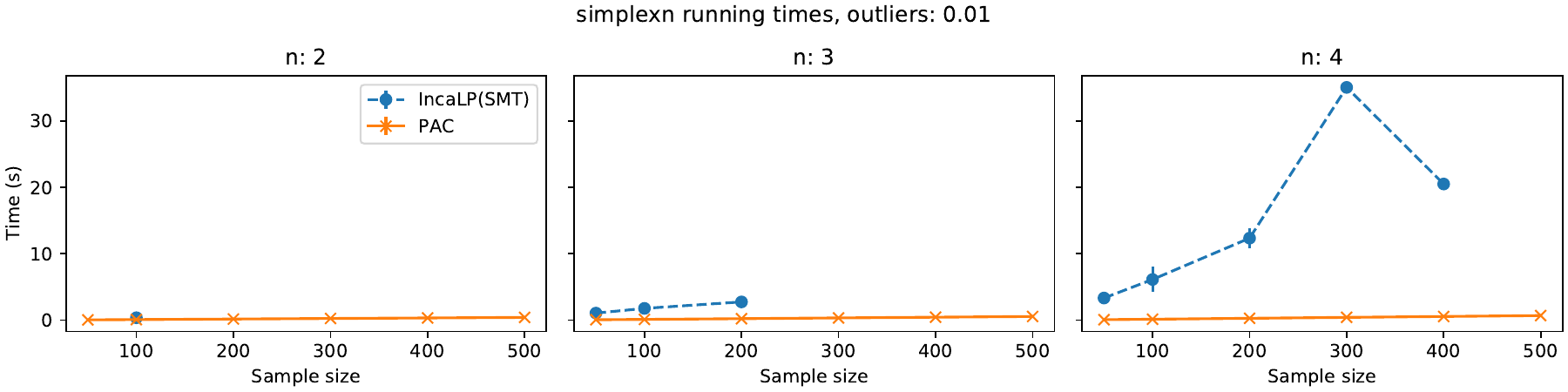}
  \end{subfigure}
  \begin{subfigure}[b]{1\textwidth}
  \includegraphics[width=1\textwidth]{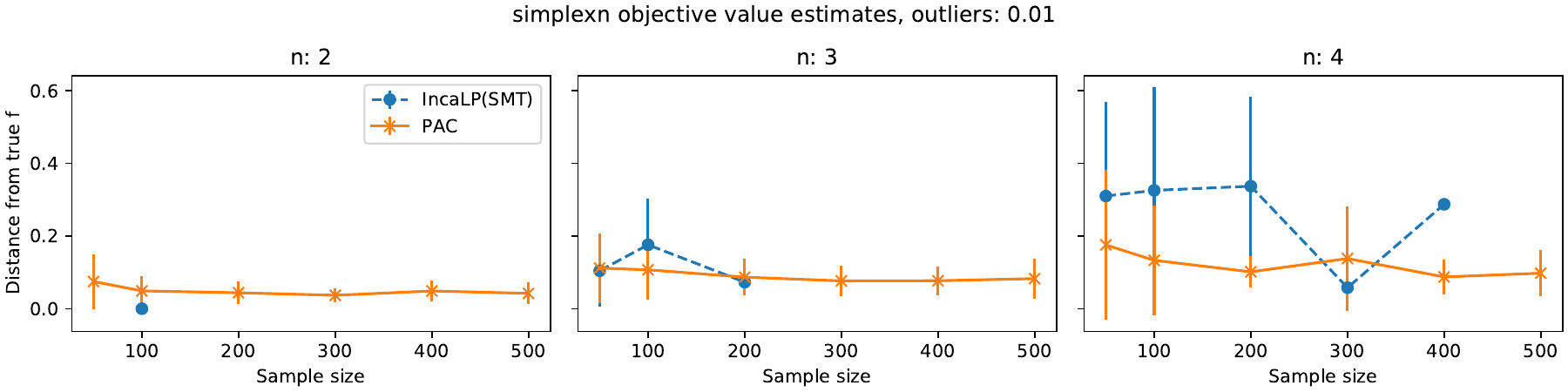}
  \end{subfigure}
  \caption{IncaLP failed to find a model 86\% of the time.}
\end{figure*}

\begin{figure*}[ht]
  \centering
  \begin{subfigure}[b]{1\textwidth}
  \includegraphics[width=1\textwidth]{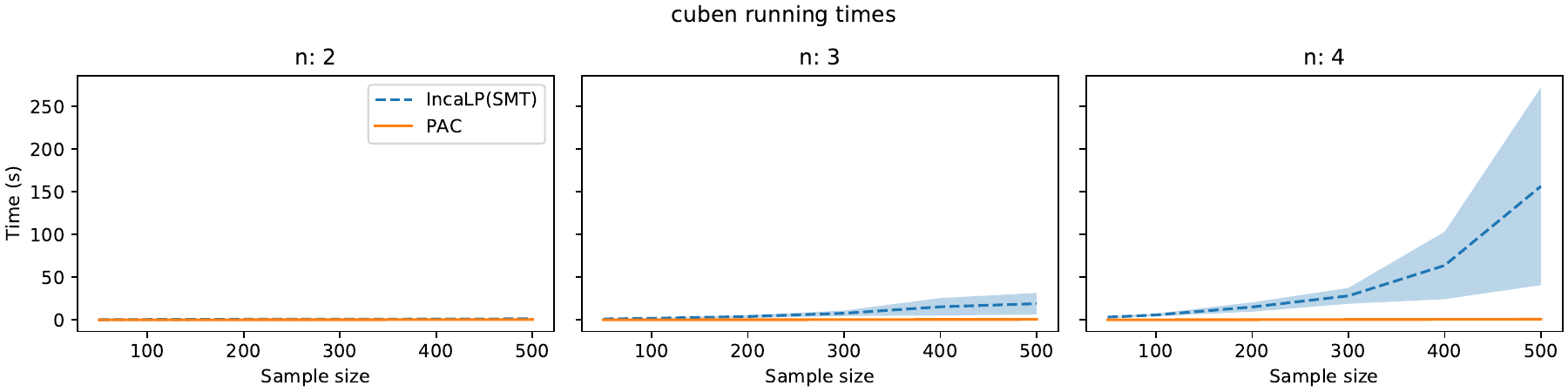}
  \end{subfigure}
  \begin{subfigure}[b]{1\textwidth}
  \includegraphics[width=1\textwidth]{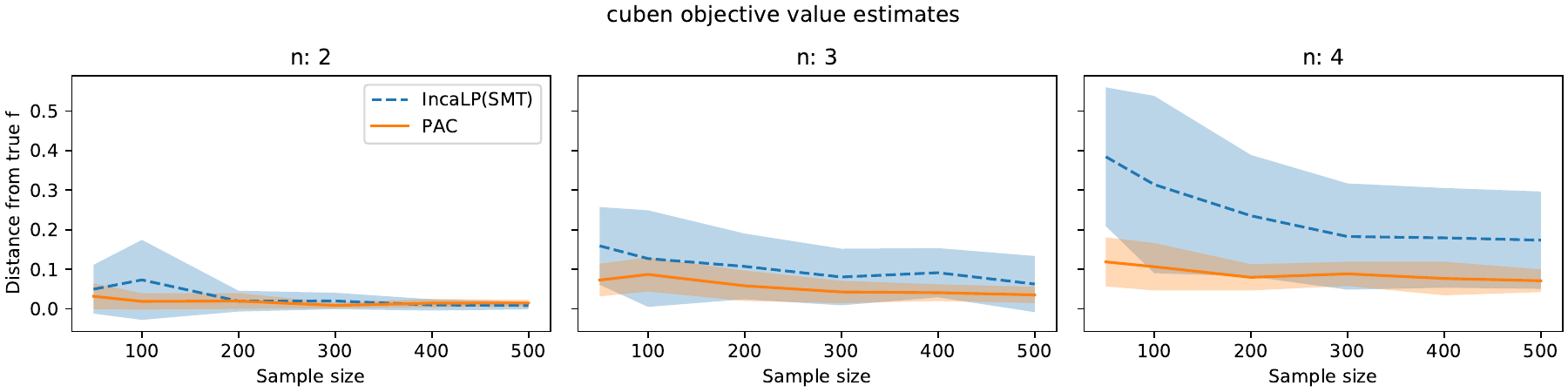}
  \end{subfigure}
  \caption{IncaLP always found a model.}
\end{figure*}

\begin{figure*}[ht]
  \centering
  \begin{subfigure}[b]{1\textwidth}
  \includegraphics[width=1\textwidth]{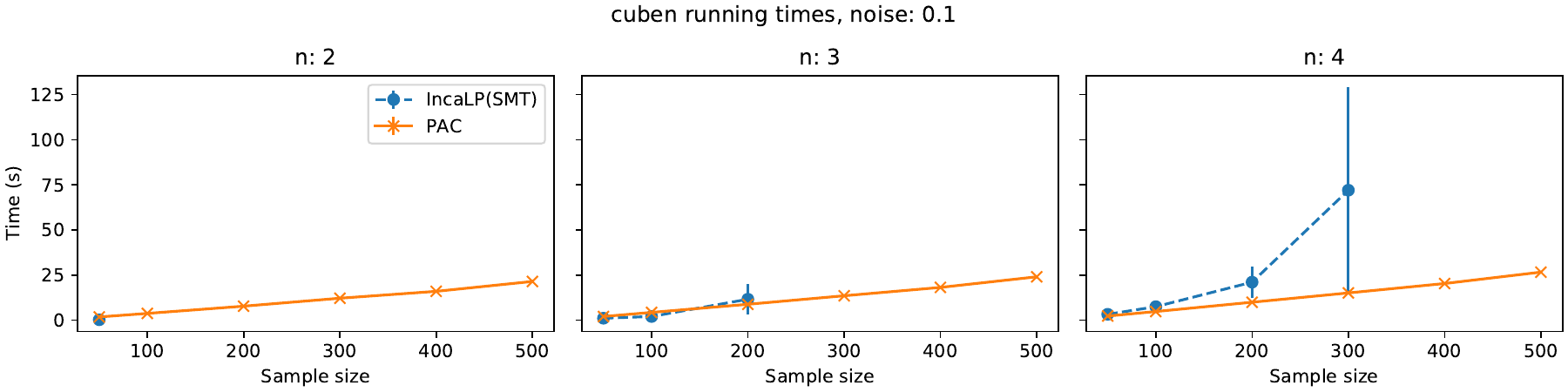}
  \end{subfigure}
  \begin{subfigure}[b]{1\textwidth}
  \includegraphics[width=1\textwidth]{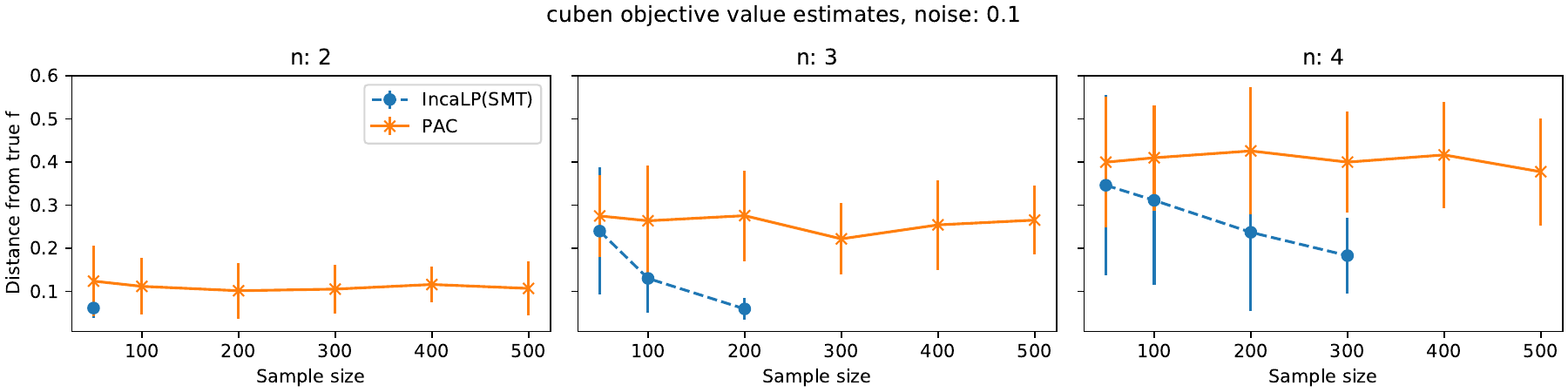}
  \end{subfigure}
  \caption{IncaLP failed to find a model 75\% of the time.}
\end{figure*}

\begin{figure*}[ht]
  \centering
  \begin{subfigure}[b]{1\textwidth}
  \includegraphics[width=1\textwidth]{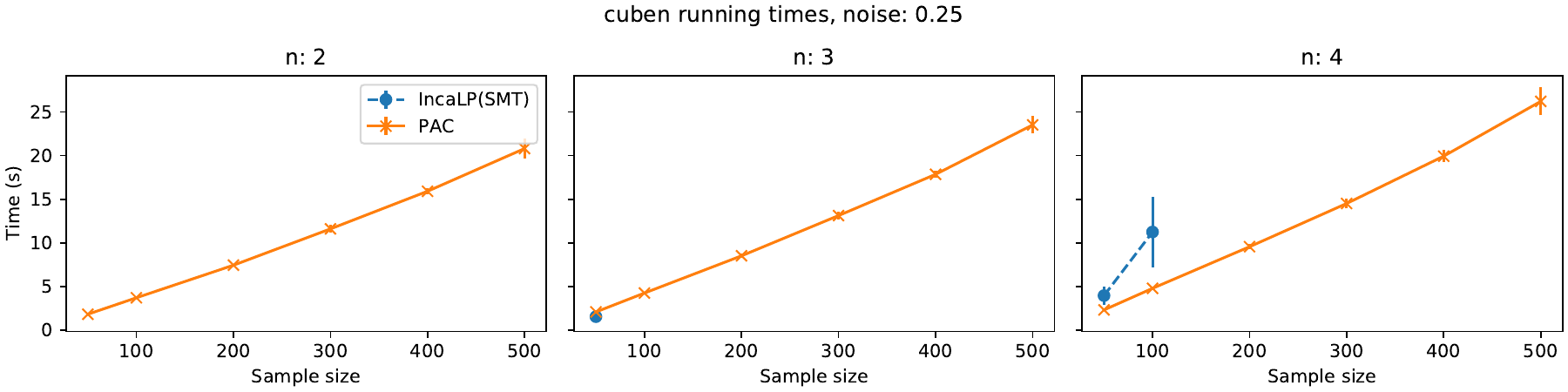}
  \end{subfigure}
  \begin{subfigure}[b]{1\textwidth}
  \includegraphics[width=1\textwidth]{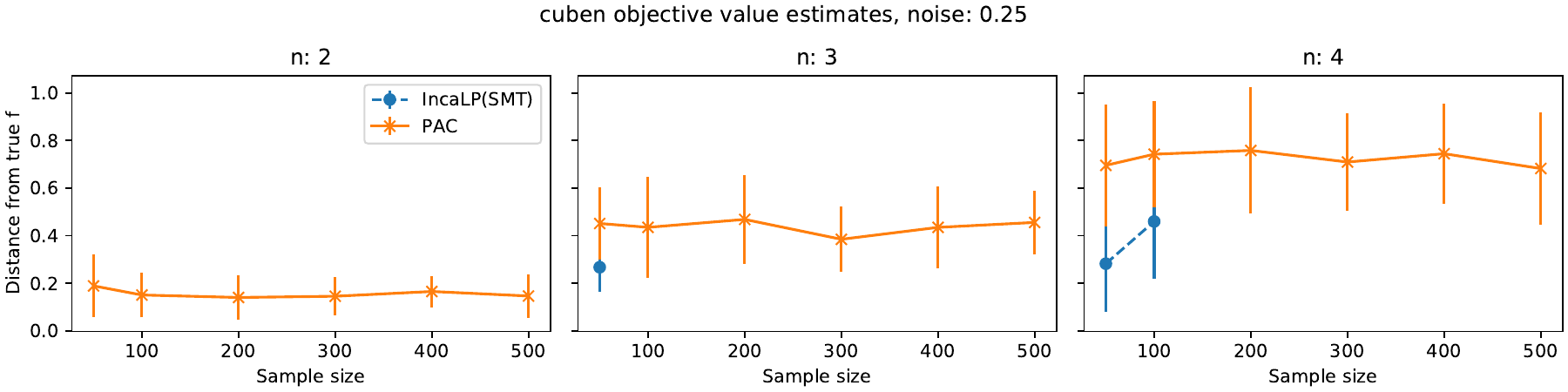}
  \end{subfigure}
  \caption{IncaLP failed to find a model 93\% of the time.}
\end{figure*}

\begin{figure*}[ht]
  \centering
  \begin{subfigure}[b]{1\textwidth}
  \includegraphics[width=1\textwidth]{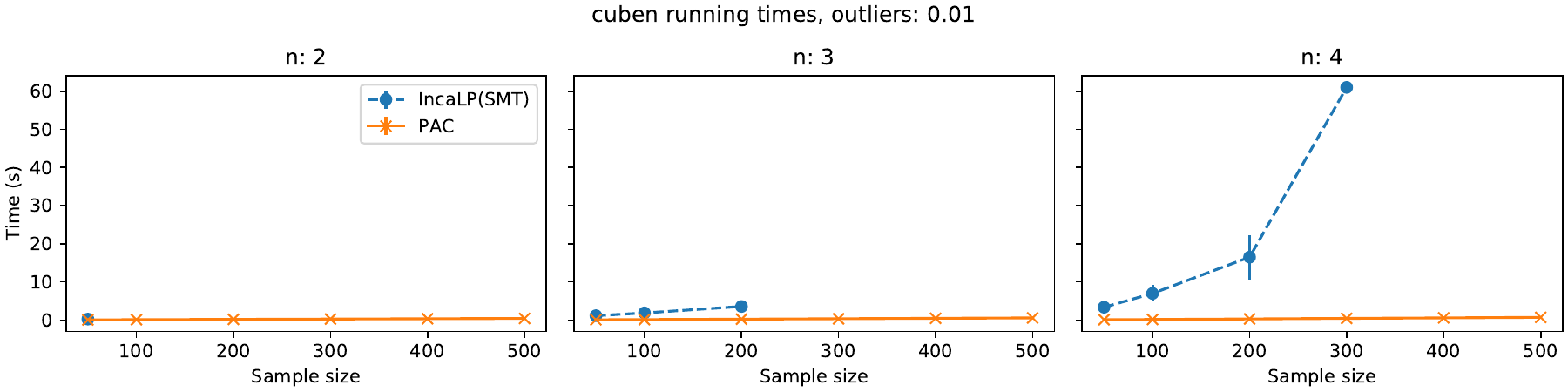}
  \end{subfigure}
  \begin{subfigure}[b]{1\textwidth}
  \includegraphics[width=1\textwidth]{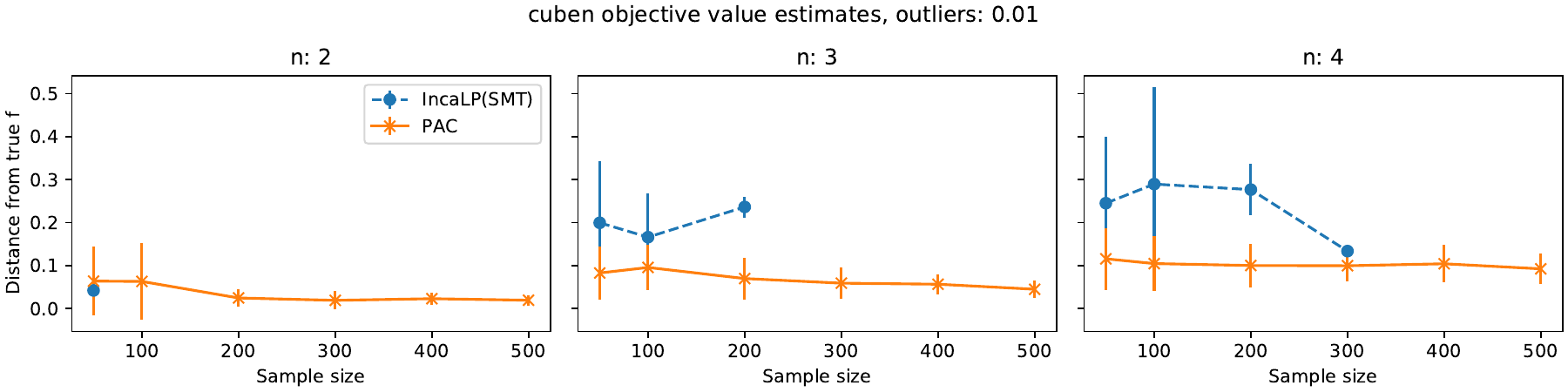}
  \end{subfigure}
  \caption{IncaLP failed to find a model 83\% of the time.}
\end{figure*}

\begin{figure*}[ht]
    \centering
    \begin{subfigure}[b]{0.49\textwidth}
    \includegraphics[width=1\textwidth]{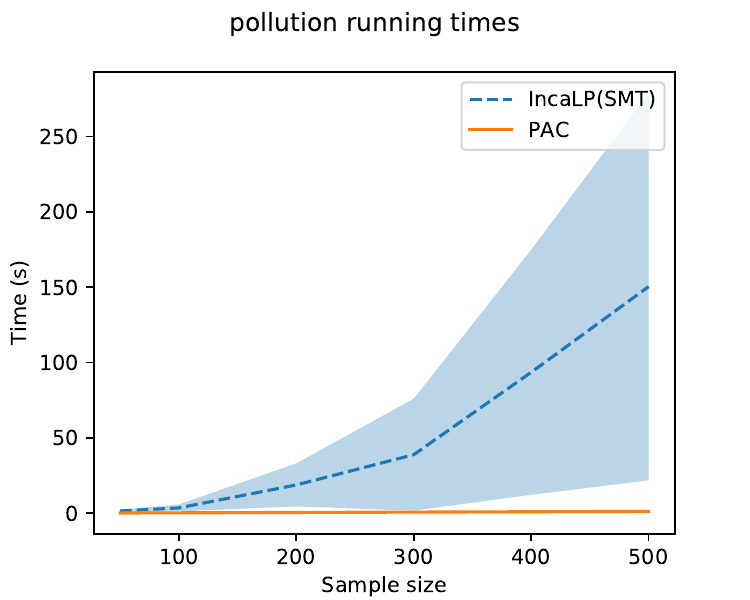}
    \end{subfigure}
    \begin{subfigure}[b]{0.49\textwidth}
    \includegraphics[width=1\textwidth]{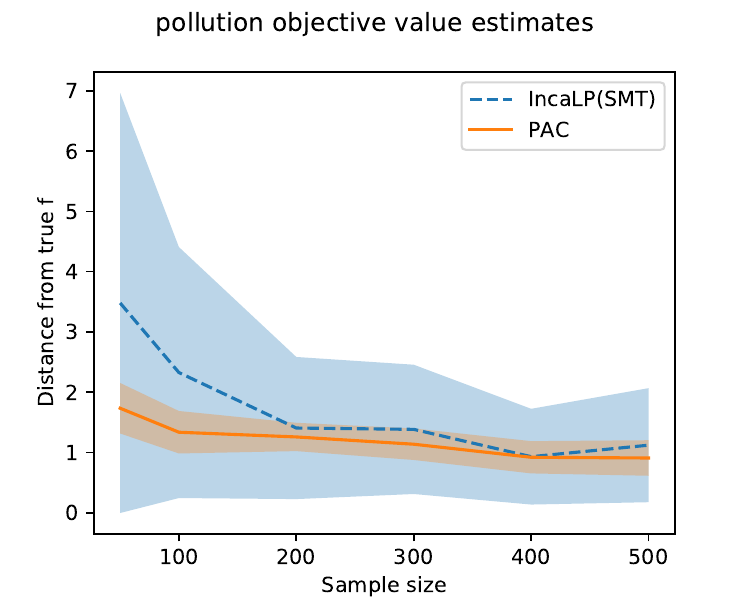}
    \end{subfigure}
    \caption{IncaLP always found a model.}
\end{figure*}

\begin{figure*}[ht]
    \centering
    \begin{subfigure}[b]{0.49\textwidth}
    \includegraphics[width=1\textwidth]{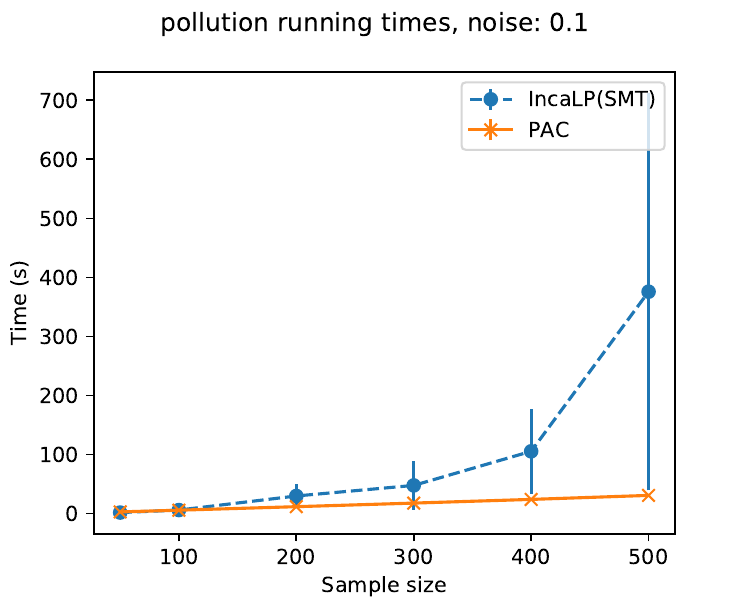}
    \end{subfigure}
    \begin{subfigure}[b]{0.49\textwidth}
    \includegraphics[width=1\textwidth]{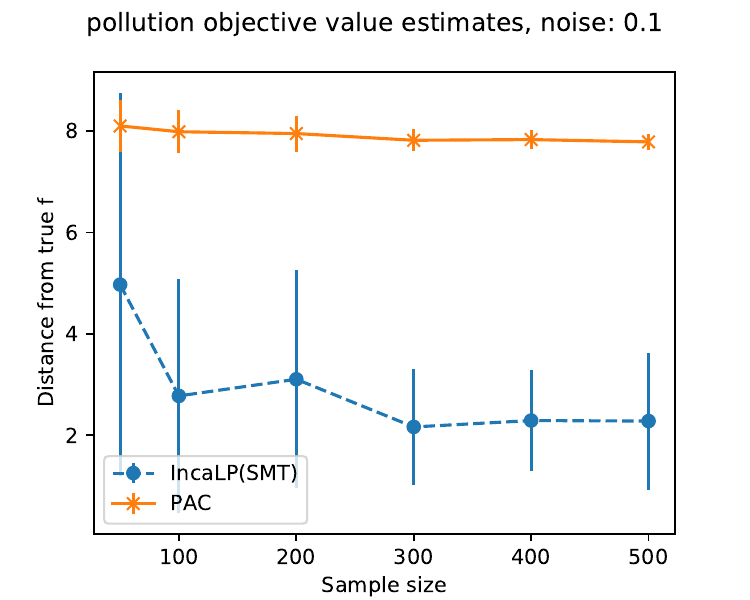}
    \end{subfigure}
    \caption{IncaLP failed to find a model 25\% of the time.}
\end{figure*}

\begin{figure*}[ht]
    \centering
    \begin{subfigure}[b]{0.49\textwidth}
    \includegraphics[width=1\textwidth]{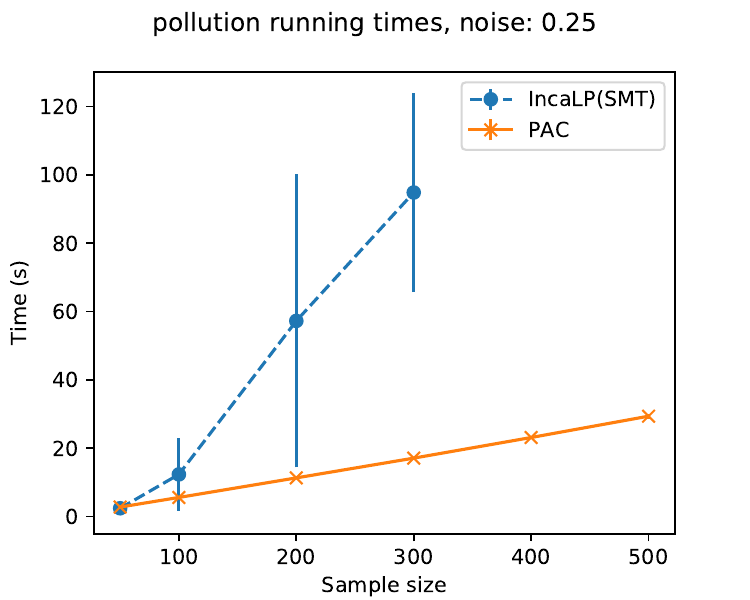}
    \end{subfigure}
    \begin{subfigure}[b]{0.49\textwidth}
    \includegraphics[width=1\textwidth]{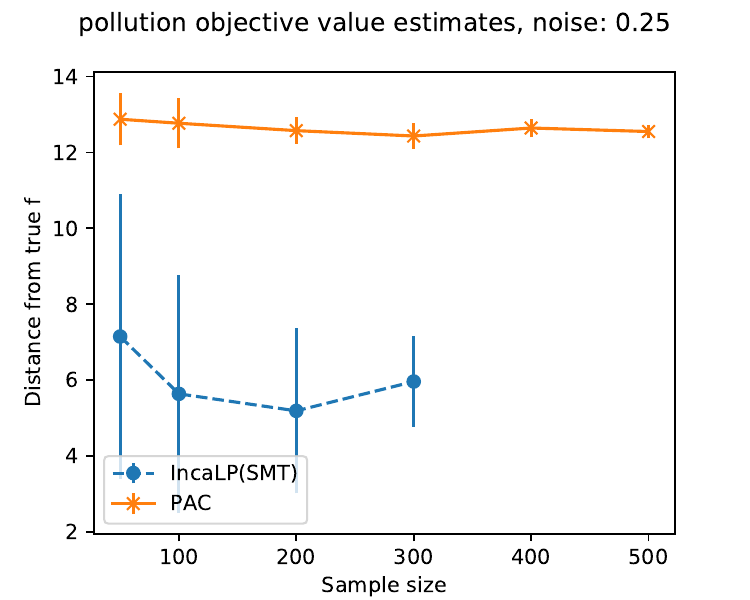}
    \end{subfigure}
    \caption{IncaLP failed to find a model 59\% of the time.}
\end{figure*}

\begin{figure*}[ht]
    \centering
    \begin{subfigure}[b]{0.49\textwidth}
    \includegraphics[width=1\textwidth]{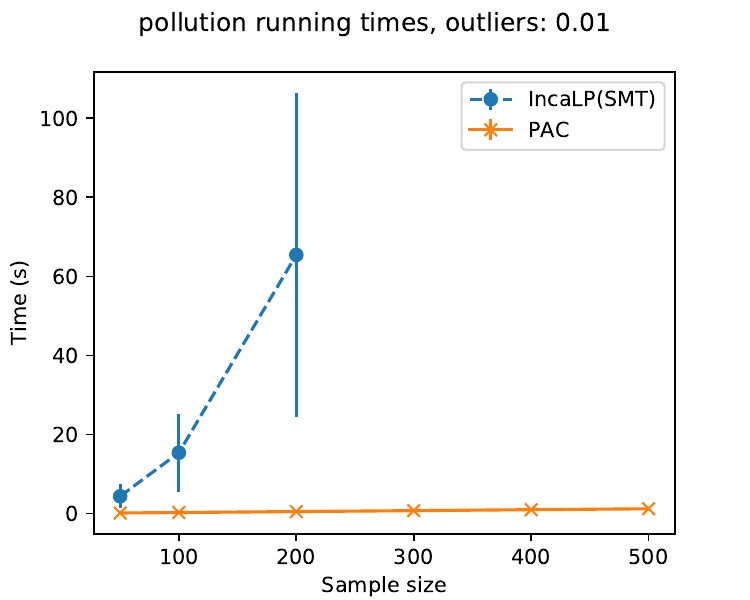}
    \end{subfigure}
    \begin{subfigure}[b]{0.49\textwidth}
    \includegraphics[width=1\textwidth]{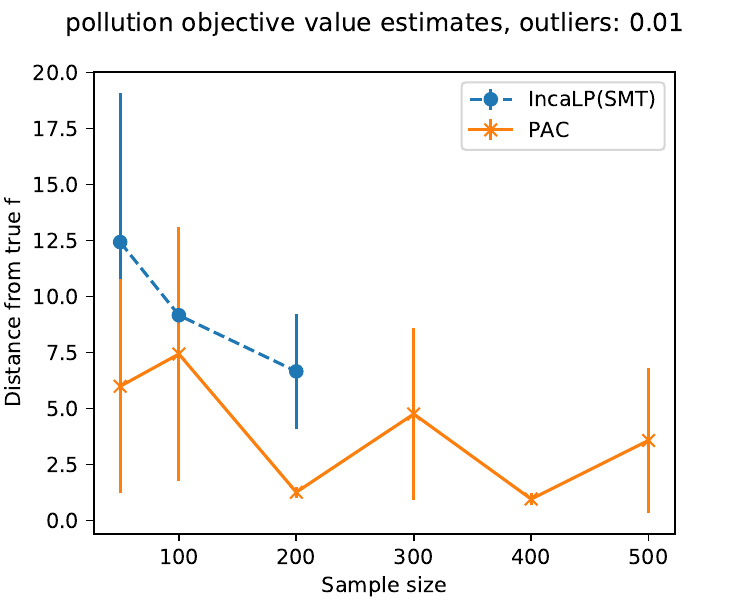}
    \end{subfigure}
    \caption{IncaLP failed to find a model 68\% of the time.}
\end{figure*}

\begin{figure*}[ht]
    \centering
    \begin{subfigure}[b]{0.49\textwidth}
    \includegraphics[width=1\textwidth]{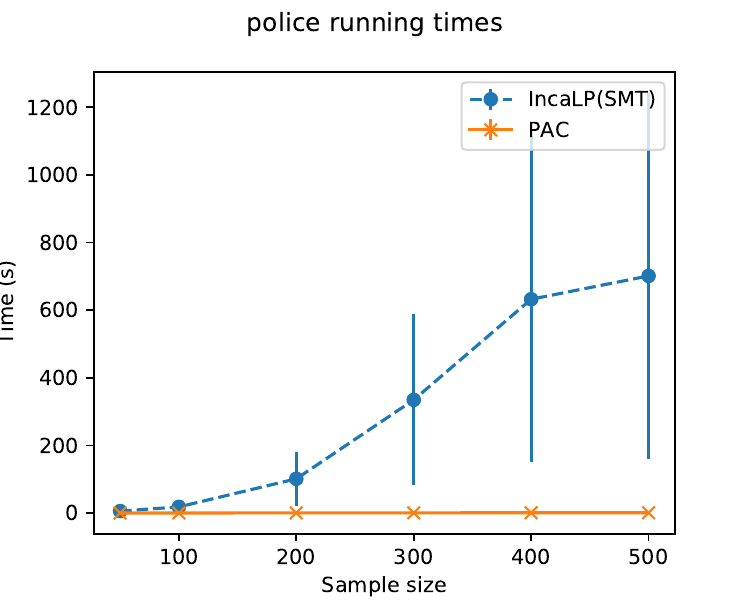}
    \end{subfigure}
    \begin{subfigure}[b]{0.49\textwidth}
    \includegraphics[width=1\textwidth]{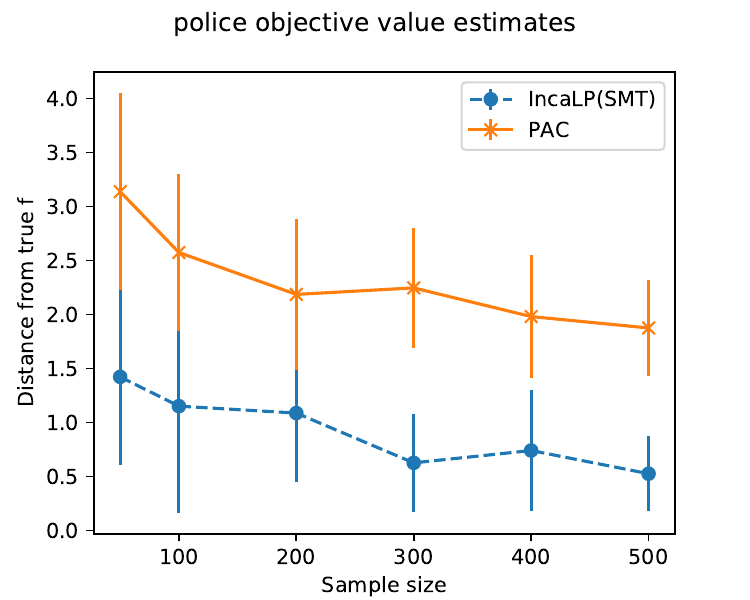}
    \end{subfigure}
    \caption{IncaLP failed to find a model 5\% of the time.}
\end{figure*}

\begin{figure*}[ht]
    \centering
    \begin{subfigure}[b]{0.49\textwidth}
    \includegraphics[width=1\textwidth]{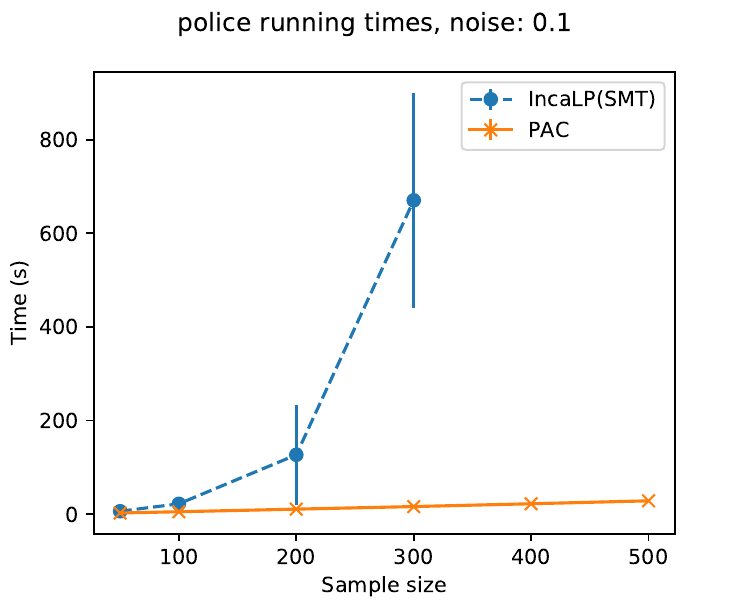}
    \end{subfigure}
    \begin{subfigure}[b]{0.49\textwidth}
    \includegraphics[width=1\textwidth]{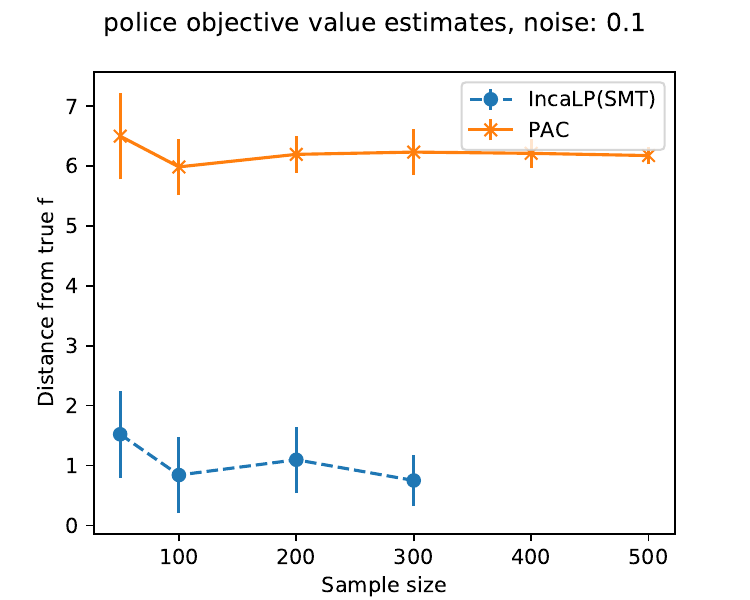}
    \end{subfigure}
    \caption{IncaLP failed to find a model 56\% of the time.}
\end{figure*}

\begin{figure*}[ht]
    \centering
    \begin{subfigure}[b]{0.49\textwidth}
    \includegraphics[width=1\textwidth]{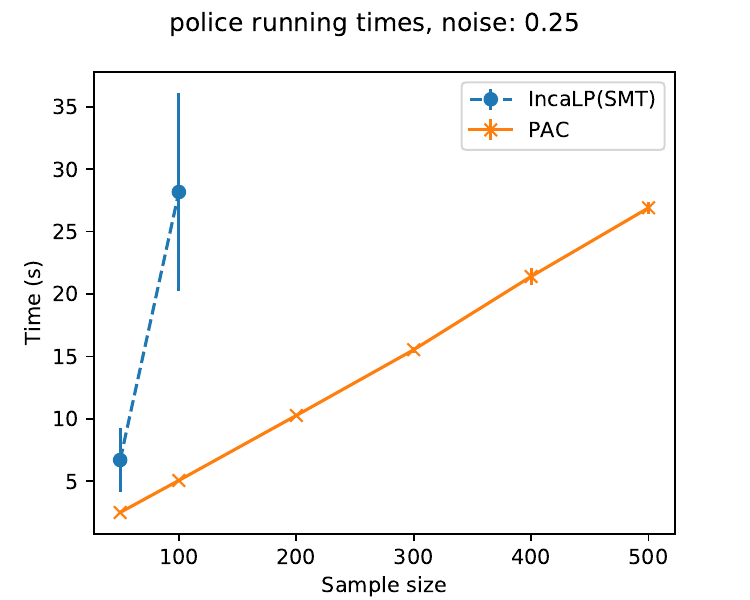}
    \end{subfigure}
    \begin{subfigure}[b]{0.49\textwidth}
    \includegraphics[width=1\textwidth]{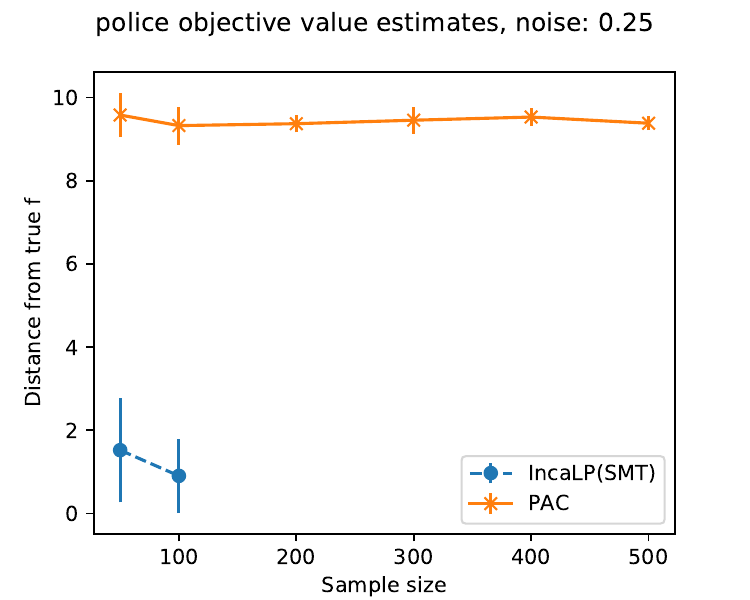}
    \end{subfigure}
    \caption{IncaLP failed to find a model 79\% of the time.}
\end{figure*}

\begin{figure*}[ht]
    \centering
    \begin{subfigure}[b]{0.49\textwidth}
    \includegraphics[width=1\textwidth]{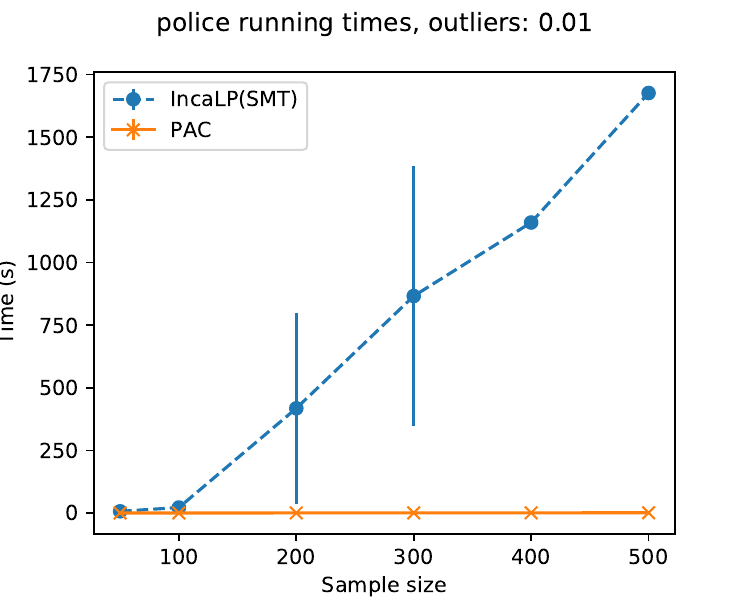}
    \end{subfigure}
    \begin{subfigure}[b]{0.49\textwidth}
    \includegraphics[width=1\textwidth]{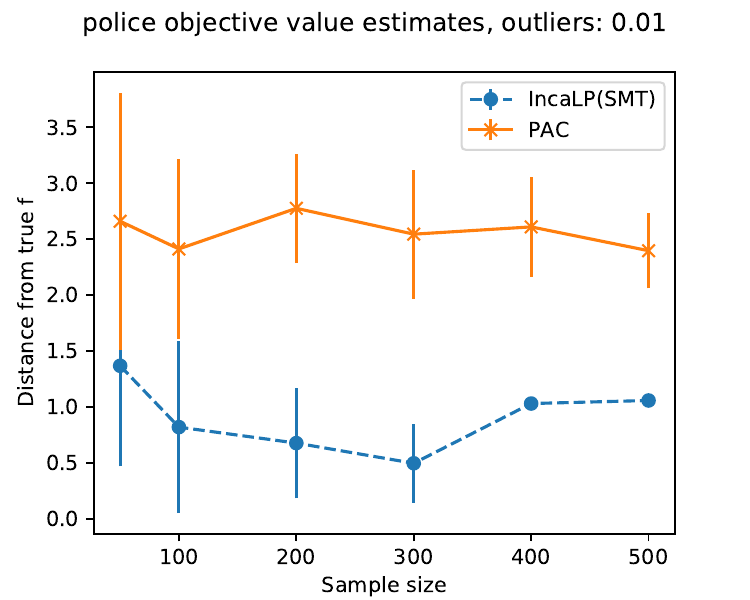}
    \end{subfigure}
    \caption{IncaLP failed to find a model 53\% of the time.}
\end{figure*}

\end{document}